\documentclass{article}

\usepackage[preprint]{corl_2026} %
\renewcommand{\paragraph}[1]{\indent\textbf{#1}}
\usepackage{xcolor}
\usepackage{wrapfig}
\usepackage{booktabs}
\usepackage{amsmath}
\usepackage{multirow}
\usepackage{subcaption}
\usepackage{graphicx}
\usepackage{makecell}
\usepackage{pgfplots}
\pgfplotsset{compat=1.18}

\newcommand{\methodname}{AVF}
\newcommand{\sysname}{EyeRobot~2.0}

\usepackage{amsmath}
\usepackage{multirow}
\usepackage{subcaption}

\title{

EyeRobot 2.0: Active Gaze for\\Precise Manipulation without Wrist Cameras

}

\author{
\normalfont
Kush Hari$^{*1}$\quad
Justin Kerr$^{*1,2}$\quad
Nidhya Shivakumar$^{1}$\quad
Samarth Mahapatra$^{1}$\quad
Carmelo Sferrazza$^{2}$\\
Jiahui Lei$^{1}$\quad
Jitendra Malik$^{1,2}$\quad
C. Karen Liu$^{2,3}$\quad
Ken Goldberg$^{\dagger 1}$\quad
Angjoo Kanazawa$^{\dagger 1,2}$
\\
\vspace{-0.7em}
\\
\url{https://eyerobot2.github.io}
\\
\vspace{-0.7em}
\\
$^{1}$UC Berkeley\quad
$^{2}$Amazon FAR\quad
$^{3}$Stanford University\quad
\\
{\small $^*$Equal contribution, random order\quad $^\dagger$Equal advising}
}

\begin{document}
\maketitle

\begin{figure}[h]
    \vspace{-5mm}
    \centering
    \includegraphics[width=\linewidth]{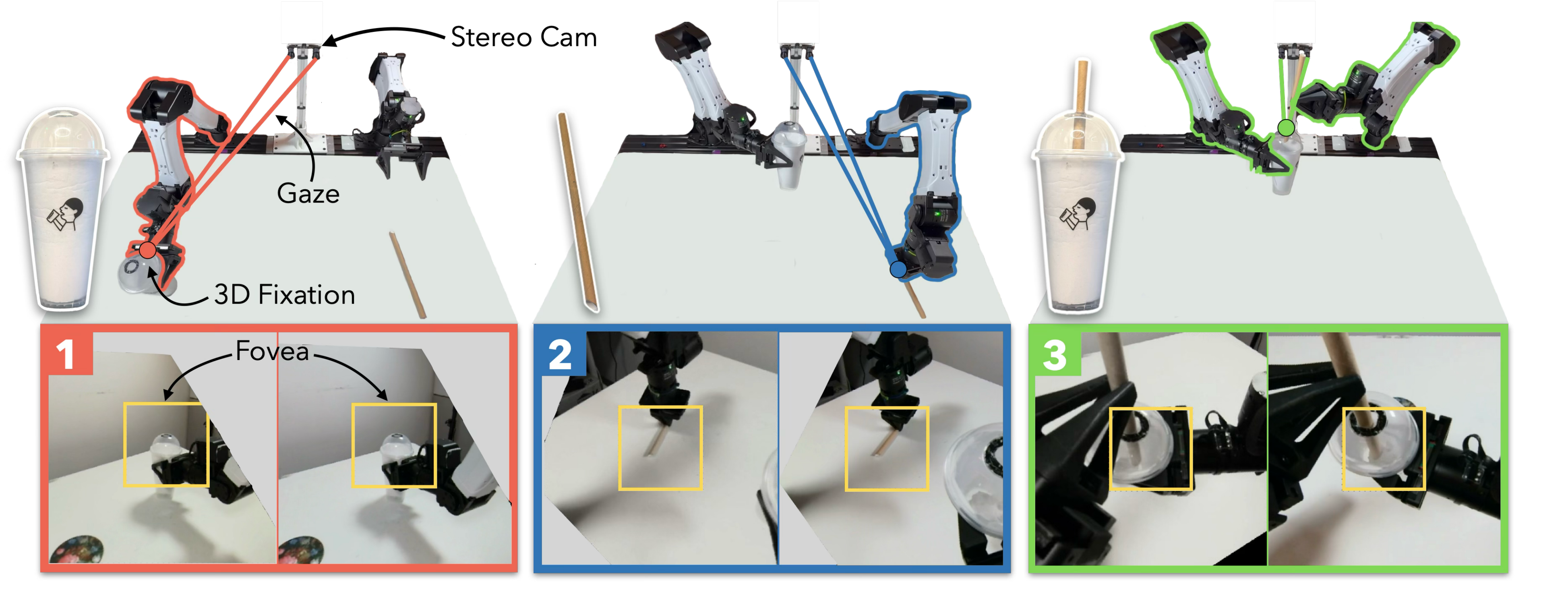}
    \caption{\textbf{EyeRobot 2.0} enables fine-grained bimanual manipulation like inserting a straw into a boba cup using only a fixed stereo camera. It coordinates a sequence of gaze targets, for example here first at the boba cup, then the straw, then the insertion point. Gaze targets are output from a semantic object selector, which conditions a low-level gaze policy to servo precisely. Its design draws inspiration from human foveated vision by leveraging multi-resolution image crops to provide high detail where the policy gazes.}
    \label{fig:teaser}
\end{figure}
\vspace{-1em}
\begin{abstract}
Inspired by human vision, we introduce a framework using active gaze to enable fine-grained bimanual manipulation with only a single stereo camera. \sysname{} physically attends to a 3D \textit{fixation point} in the scene by swiveling two eye viewpoints to center their gaze on it. The resulting images are processed foveally by allocating more visual tokens to the image centers, focusing computation on task-relevant features. Such Active Visual Fixation (\methodname{}) requires carefully coordinated gaze during task execution, which we accomplish hierarchically by first training a low-level gaze servoing policy conditioned on a goal object, then training a target selector which emits fixation goals based on task progress. Both modules are trained with RL on real-world data: the first is trained with a dense geometric reward and the second co-trains with the BC gripper policy which allows it to discover fixation sequences that can resemble a human's fixation sequence while performing the task (Fig.~\ref{fig:teaser}). \sysname{} further takes advantage of fixation by canonicalizing gripper proprioception and actions into a rotating fixation-relative $\text{SE}(3)$ frame, which compacts the size of the action distribution to learn.
We collect teleoperation data for 7 real-world and 6 simulated tasks, and conduct over 1000 physical and 1800 simulated robot trials comparing \sysname{} against passive stereo and ego $+$ wrist camera policies trained on the same data. Removing wrist cameras is costly for standard policies: with only passive stereo, real-world success drops from $52\%$ to $27\%$. \sysname{} closes this gap without any wrist cameras, outperforming passive stereo by $40\%$ in real and $20\%$ in sim. It matches ego $+$ wrist policies when their wrist views are clear ($69\%$ vs.\ $64\%$), and more than doubles their success when grasped objects occlude the wrist cameras ($48\%$ vs.\ $22\%$). These results suggest active gaze as a single-camera alternative to wrist cameras for fine-grained manipulation.

\end{abstract}

\section{Introduction}
\label{sec:introduction}
Consider the task of picking up a cup with a lid, grasping a straw, and inserting the straw into a small opening in the lid (Fig.~\ref{fig:teaser}). As humans carry out this task, our eyes flit around the scene, filtering information with coordinated look-ahead movements that anticipate hand motion~\cite{LAND20013559}. This task-relevant visual sampling, called \textit{fixation}, directs high-resolution foveal vision to relevant parts of the visual field. Fixation also provides 3D anchor points for manipulation, simplifying the challenge of spatially reasoning about the scene~\cite{Feldman_1985} and implicitly breaking larger tasks into concrete sub-goals as gaze switches between targets.

We hypothesize that this behavior could also benefit robot manipulation.
Fixating on a dynamic task-relevant 3D point acts as a form of physical attention which
centers input observations, compacts the data distribution, filters irrelevant visual regions, and allocates additional resolution where it matters for fine-grained tasks without the cost of processing the full panorama at high resolution.
Some of these benefits can be offered by wrist-mounted cameras, which are used by nearly all existing bimanual manipulation systems~\cite{zhao2023learning,chi2024universal,black2410pi0,fu2024mobile}. However, colocating vision with the gripper ties precise visual input to the manipulator, which is limiting in tool use, whole-body manipulation, or when grasped objects occlude the camera.
In addition, these systems constrain manipulator design because of their direct mounting to the gripper, which adds complexity, often occludes the ego camera, introduces motion blur, and prevents sleeker gripper designs. A purely ego-centric behavior cloning approach has the potential to ease constraints on end effector design, make better use of ego-centric human data, and improve tasks which benefit from vision-manipulator decoupling like fast motion, reaching into tight spaces, or searching large spaces~\cite{kerr2025eye}.

We present \sysname{}, a robot learning framework which introduces Active Visual Fixation (\methodname{}). Using input \textit{only} from one fixed stereo camera, \sysname{} can carry out precise manipulation tasks like inserting a straw, capping a marker, zipping a bag, or extracting a wrench from a toolkit, and multi-stage tasks like opening a toaster oven, pulling the tray out, extracting a pan, pushing the tray back, and closing the door again.

\sysname{} couples a fixation policy with a fixation-centric gripper policy. The fixation policy has two modules: a low-level gaze policy that coordinates two eye viewpoints to rapidly find and track 3D objects in the scene, and a high-level target selector to choose \textit{what} object to look at and \textit{when} to look at it during the task. Visual inputs are represented as foveated stereo multi-crops centered at the fixation point, which allocate more resolution towards the direction of interest. This physical attention not only increases visual acuity, but also reduces computation allocated to distractors.
In addition, \sysname{} trains a fixation-centric gripper policy via behavior cloning, by representing SE(3) proprioception and action chunks canonicalized into the rotated fixation frame. This effectively constrains the input and output space for the gripper model with respect to task-relevant local regions. The gripper policy additionally receives the distance of the 3D fixation point as an observation, which acts as an anchor point for the policy to understand depth.
Both the gaze policy and target selector are trained with reinforcement learning (RL), with gaze emerging from a BC-RL loop as in~\citet{kerr2025eye}.

We evaluate \sysname{} on 7 physical tasks and 6 simulated MuJoCo tasks, collecting data and training single-task policies from scratch. Across over \textit{1000 physical trials}, it outperforms static stereo baselines by 40\%, and by 20\% on simulated pick-place tasks, with a larger performance improvement on fine-precision tasks. Compared to the widely adopted SOTA ego + wrist setup, \sysname{} \textit{without wrist cameras} matches task performance in physical trials, and on tasks with heavy occlusion of wrist cameras maintains strong performance while wrist baselines regress to static stereo performance. In ablations we find that fixation-centric actions account for a significant portion of \sysname{}'s improvement, causing a 20\% performance boost across sim tasks, and foveation contributes strongly to real-world success on fine-grained tasks, boosting by 20\% overall.

\begin{figure}[t]
    \centering
    \includegraphics[width=\linewidth]{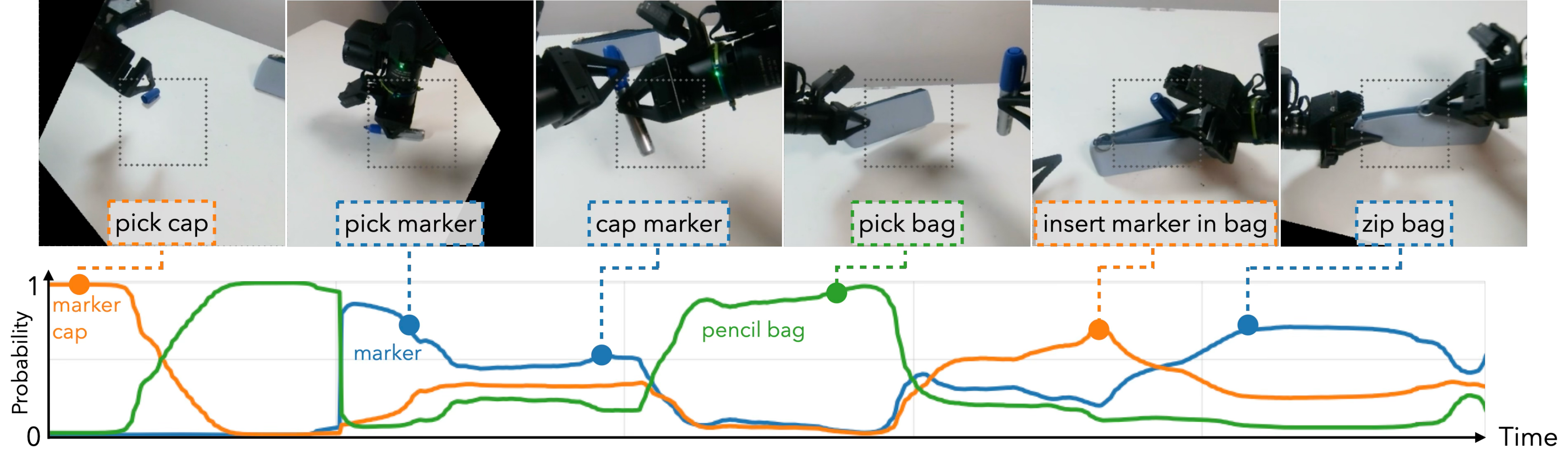}
    \caption{\textbf{Gaze target outputs} over an autonomous physical trial which picks up a marker cap and body, caps it, repositions a bag, inserts the marker, and zips the bag. The target selector outputs goals that are relevant to the current stage of the task: for example the policy looks at the marker when capping or the bag when repositioning. Wrist cameras are present but unused.}
    \label{fig:probabilities}
    \vspace{-1em}
\end{figure}

\section{Related Work}
\label{sec:related_work}

\subsection{Learning manipulation from human demonstrations}
Behavior Cloning (BC) allows robots to learn complex behaviors from teleoperated demonstrations without requiring handcrafted motion primitives and reward functions~\cite{pomerleau1988alvinn,hussein2017imitation,ravichandar2020recent,chi2025diffusion}. BC has been extended to work in complex settings including mobile manipulation \cite{du2022bayesian,fu2024mobile} and bimanual manipulation~\cite{zhao2023learning,chi2024universal}, and several recent efforts have been directed towards scaling teleoperation datasets~\cite{brohan2022rt,khazatsky2024droid,black2410pi0, open_x_embodiment_rt_x_2023,zhao2024alohaunleashedsimplerecipe}. Another enticing direction for scaling is learning from ego-centric human data, either by instrumenting humans or scraping internet videos~\cite{grauman2022ego4dworld3000hours,grauman2024ego,lv2024ariaeverydayactivitiesdataset,liu2022hoi4d,gao2026dreamdojogeneralistrobotworld}. However, learning from human videos presents a significant domain gap between ego-centric sensors in first-person datasets and the wrist cameras that our robot policies so frequently rely on for fine-grained performance. \sysname{} aims to provide a route towards lessening this gap by leveraging human-like fixation for precise bimanual manipulation. We are also inspired by task conditioning approaches in BC which condition a low-level policy on the output of a higher-level task sequencer~\cite{lynch2020learning,wang2023mimicplay,shi2024yell,rth2024arxiv}. Similarly, \sysname{} learns a target selector to control gaze, but does not require human gaze supervision and instead trains gaze with RL via a BC accuracy reward.

\subsection{Active vision for manipulation}
Active vision deliberately moves sensors to improve observability, filter information, and reduce occlusions for better environment understanding \cite{goldberg1984active,bajcsy1988active,aloimonos1988active,bajcsy2018revisiting,jayaraman2018learning,van2021active,ballard1991animate,tsotsos1992relative}. 
Previous systems have used active vision to help with tasks such as visual search \cite{zhu2017target,kolve2017ai2,ye2018active,rasouli2020attention,dass2024learninglookseekinginformation}, object detection and tracking \cite{papanikolopoulos1993visual,wilkes1994active,gould2007peripheral,jonnalagadda2021foveater}, and 3D reconstruction \cite{banta1995best,isler2016information,mendoza2020supervised,zeng2020view}.
Similarly, active vision has proven beneficial for robot manipulation with a ``robot neck'' moving the camera based on VR headset movement during task teleoperation to reveal better viewpoints~\cite{cheng2024open,chuang2025active,zeng2025activeumi,xiong2025vision,yu2025egomi,spigler2026tavis}. Active visual search can also benefit from training on ego-centric human data~\cite{li2021eye,pani2026gaze,li2026actsenseactlearning}. \sysname{} in contrast does not attempt to address viewpoint selection which has been explored in prior works, but rather investigates how coordinated fine-grained eye movements can boost performance \textit{even with a well-positioned viewpoint}.

Related to learning more precise gaze, some papers collect human eye tracking data during teleoperation to predict a gaze point as an intermediate output to condition a gripper policy with~\cite{chuang2025look,li2026gazevla} or crop a high-resolution finger-tip image~\cite{kim2024multi}. Others~\cite{lee2025gaze} learn gaze in simulation through on-policy task-based rewards; however, these methods struggle with sim-to-real transfer. EyeRobot~\cite{kerr2025eye} overcomes the need for gaze demonstrations by training a gaze policy on top of real-world behavior cloning data from $360^\circ$ video by co-training the eye and arm policies with an action feedback loop for reward. Building on this, \sysname{} uses a similar co-training loop but tackles much more fine-grained manipulation tasks by utilizing active stereo fixation, a fixation-centric action representation, and a hierarchical target selector to precisely coordinate gaze throughout execution.

\section{Manipulation with Active Visual Fixation}
\textbf{Problem Statement}: 
We study visuomotor policy learning from human teleoperation demonstrations for bimanual manipulation with parallel-jaw grippers.
Given a set of demonstrations specifying a task, our goal is to train task-specific policies that can reproduce the demonstrated behavior with bounded variations in pose. 
In particular, \sysname{} considers an ego-centric setting in which observations are provided solely through an approximately head-mounted stereo camera. The policy is evaluated using task-specific success criteria.

\textbf{Overview}: \methodname{} trains separate \textit{gripper} and \textit{fixation} policies that work together to achieve a task. The two eyes' viewpoints actively pivot to converge on a shared 3D fixation point, whose location is updated at each timestep by the fixation policy. The resulting fixation-centered observations are fed into the downstream gripper policy, which produces action chunks for robot execution. These SE(3) action chunks are canonicalized into a dynamically shifting fixation frame pointing towards the current fixation point (Fig.~\ref{fig:method-arch}). Each policy uses a foveated transformer-decoder architecture similar to \cite{kerr2025eye}, which we fully describe in the Appendix.

\begin{figure}[t]
    \centering
    \includegraphics[width=0.95\linewidth]{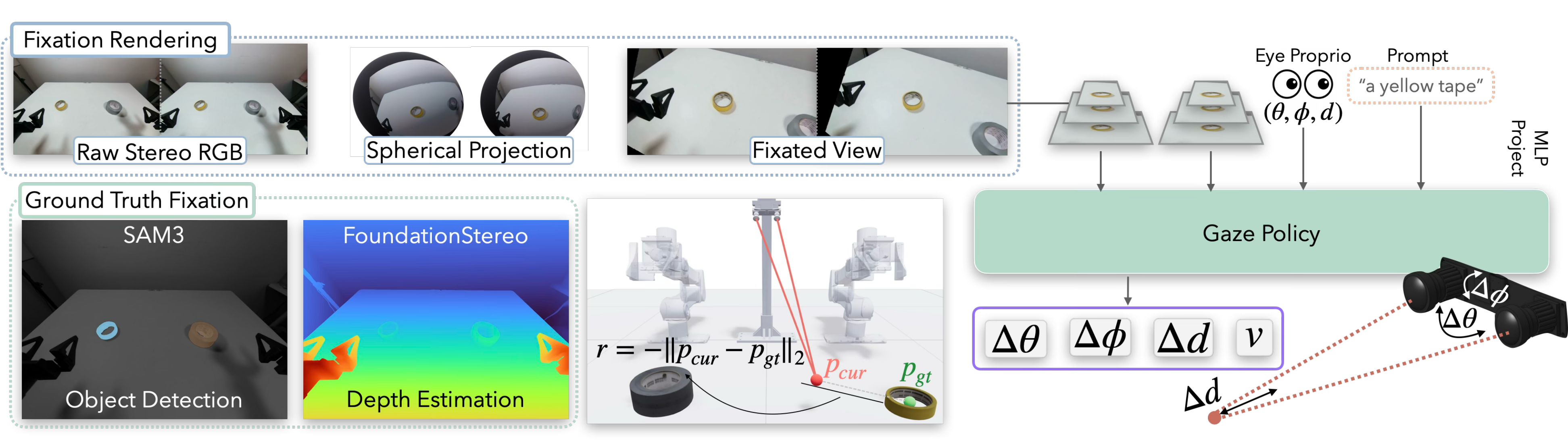}
    \caption{\textbf{Learning gaze control}. \sysname{} trains a goal-conditioned gaze policy to continuously servo two eyes' gaze in 3D. We train this behavior via RL, with a spatial reward derived from detecting the object and estimating its 3D centroid with stereo depth.}
    \label{fig:method-3D}
    \vspace{-1em}
\end{figure}

\subsection{Learning Gaze Patterns with RL}

We train \methodname{} in two phases. First, we train a goal-conditioned fixation policy capable of fixating on any task-relevant objects. Next, we train a hierarchical target selection policy that sequences fixations to optimize task performance. 
In the first phase, the fixation policy is trained with RL using a geometric reward from detected object positions, enabling accurate 3D tracking of prompted objects. In the second phase, the target selector is trained with RL in a BC-RL loop as in EyeRobot~\cite{kerr2025eye}: a gripper BC policy is co-trained on the resulting fixated views, and the selector is rewarded for fixations that maximize the BC policy's action prediction accuracy. This hierarchical formulation allows task-dependent gaze behavior to naturally arise without the need for manual human specification.

To learn gaze, we need a way to counterfactually render it on top of existing robot demonstrations that were collected before any gaze policy exists. We therefore need to render what the robot \textit{would have seen} had it looked somewhere. EyeRobot~\cite{kerr2025eye} addressed this by collecting demonstrations with a $360^\circ$ camera and rendering novel views from it. We transfer this idea to stereo, using a calibrated stereo camera to synthesize fixation: rotating a camera about its optical center only requires warping its image, without any 3D reconstruction. This lets \methodname{} use the same stereo camera for data collection, training, and deployment.
We simulate fixation by rendering viewpoints from a stereo camera to replicate the appearance of two eyes converging their gaze on a common 3D point in the scene.

\paragraph{Goal-Conditioned Stereo Fixation}
First, we train a standalone goal-conditioned fixation policy. This policy takes as input a language prompt specifying the goal, the current fixated observations, and the current fixation point in polar coordinates. Though in principle this language prompt could be anything, in practice we limit it to a task-specific annotated set of target objects which the robot interacts with in the scene. 
The gaze policy outputs local fixation deltas for azimuth and elevation $\Delta\theta, \Delta\phi$, and fixation distance $\Delta d$ represented as a discretized distribution. 

We train this policy with PPO~\cite{schulmanppo} on rollouts from randomly selected stereo still-frames from our teleop demonstration set.
To compute a goal-conditioned reward $r$ for each rollout, we first detect candidate objects by querying SAM3~\cite{carion2025sam3segmentconcepts} with the task's language prompts, then randomly select one detected prompt. To localize the object in 3D, we extract depth from the calibrated stereo pair with FoundationStereo~\cite{wen2025stereo} and set the ground truth position to be the 3D centroid of depth intersected with the object mask, $p_{gt}$. At each step of RL, the policy is conditioned on the current prompt and rewarded based on the current fixation point $p_{cur}$ with $r=-||p_{cur}-p_{gt}||_2$. The result is a lightweight policy that can rapidly fixate on all task-relevant objects at 30\,Hz. 

\begin{figure}[t]
    \centering
    \includegraphics[width=0.95\linewidth]{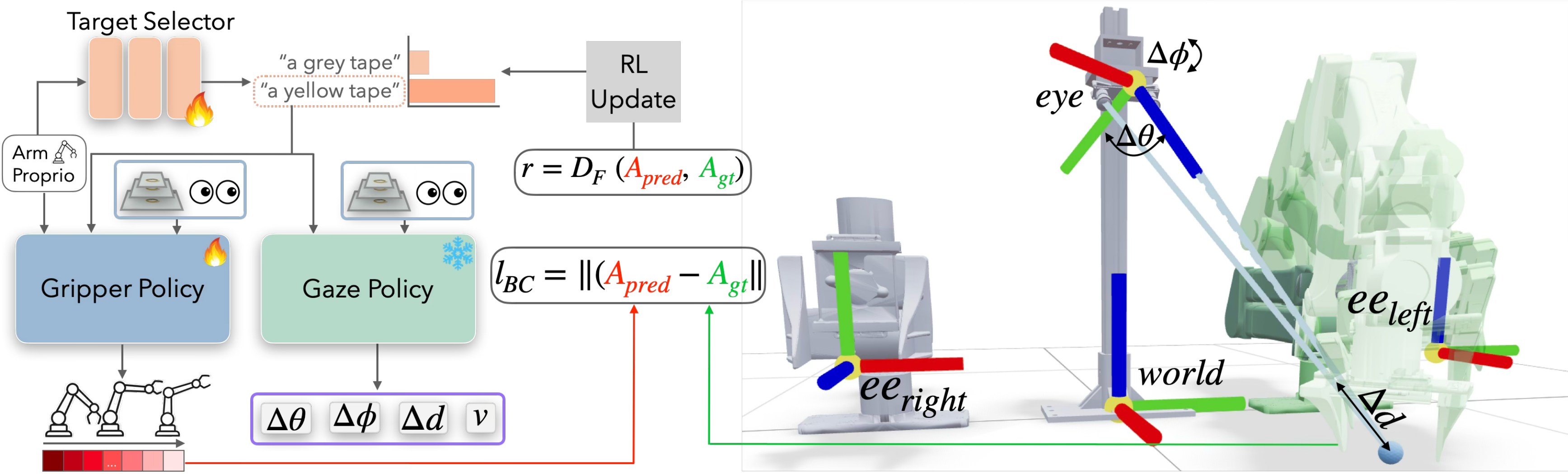}
    \caption{\textbf{Sequencing fixation for manipulation.} (\textit{Left}) \sysname{} trains a lightweight per-task target selector to switch between objects in the scene. 
    This target selector is trained with RL to maximize the prediction accuracy of a gripper BC policy, which is co-trained from scratch at the same time on viewpoints generated from the fixation rollout. Action chunks are supervised with teleop demos, and negative prediction error is supplied as the reward for the selector to optimize. (\textit{Right}) Action chunks are represented as SE(3) chunks in the eye frame, aligned with the current fixation point.}
    \label{fig:method-arch}
    \vspace{-1em}
\end{figure}

\paragraph{Learning to Sequence Fixations via a BC-RL Loop}
While the first phase enables low-level gaze behavior capable of fixating on any object, we now address the high-level decision of \textbf{where} and \textbf{when} to focus. 
As in EyeRobot~\cite{kerr2025eye}, to circumvent the burden of hand-annotated task stages, we learn this decision process via RL by jointly training gaze alongside the BC policy itself. This naturally encourages gaze sequences that optimize task performance.
To do this, we freeze the goal-conditioned fixation policy, and train a lightweight \textit{target selector} with RL, which is conditioned on robot proprioception and outputs a categorical distribution over per-task prompts (Fig.~\ref{fig:method-arch}, top left). The resulting sampled prompt is used to condition the fixation policy, whose behavior is used during PPO rollout as observations to compute gripper action chunk predictions, which are then used to compute reward for the target selector to optimize (Fig.~\ref{fig:method-arch}, middle). During training, the target selector emits a new target once every 15 frames to allow gaze to settle before sampling a new target.

During train rollouts we uniformly sample a start frame from the demonstration, with special 10\% weighting given to the first frame in the demo which often contains the most critical decision for gaze (what to reach for first). 
The first 30 frames of the agent's observations operate on a paused frame to allow gaze to settle before being forced to act. 
We adopt a similar per-step reward as \cite{kerr2025eye}, which extracts a task performance proxy from the BC policy's prediction error. Given predicted and ground truth bimanual action chunks transformed into the same frame $A_{pred},~A_{gt}$, we first identify an ``active arm'' whose end effector moves a greater distance in $A_{gt}$, then compute the arc-length Fréchet distance between $A_{gt}$ and $A_{pred}$ for this arm. The negative of this distance is used as reward to optimize the target selector, encouraging fixation which aids BC prediction accuracy. Limiting reward computation to the active arm avoids sensitivity to spurious correlations in bimanual demonstrations, and Fréchet distance provides a more stable velocity-invariant metric than L2 error.

\subsection{Fixation-Centric Action Chunking}
Knowing where to focus not only reduces perceptual complexity but also improves action prediction. Specifically, we represent the robot's proprioception and action chunks relative to the fixation frame.
Such a representation standardizes the input about the fixation direction, making the training distribution more compact. 
Model input proprioception and actions are represented as SE(3) end effector poses transformed into the fixation frame, defined as the cyclopean (virtual center-eye) rotation towards the fixation point in 3D. 
This canonicalized action representation moves dynamically with fixation, which results in a significant performance improvement over a fixed world frame.
We jointly train the BC policy alongside gaze post-training on perspectives sampled from the fixations induced by RL policy rollout.
We maintain a ring buffer of the past 100 rollouts seen, and sample BC train batches independently from this buffer to maintain diversity across demonstrations. The BC architecture is a transformer decoder resembling that of \cite{kerr2025eye}; see the Appendix for architecture and hyperparameter details.

\begin{figure*}[t]
    \centering
    \includegraphics[width=\linewidth]{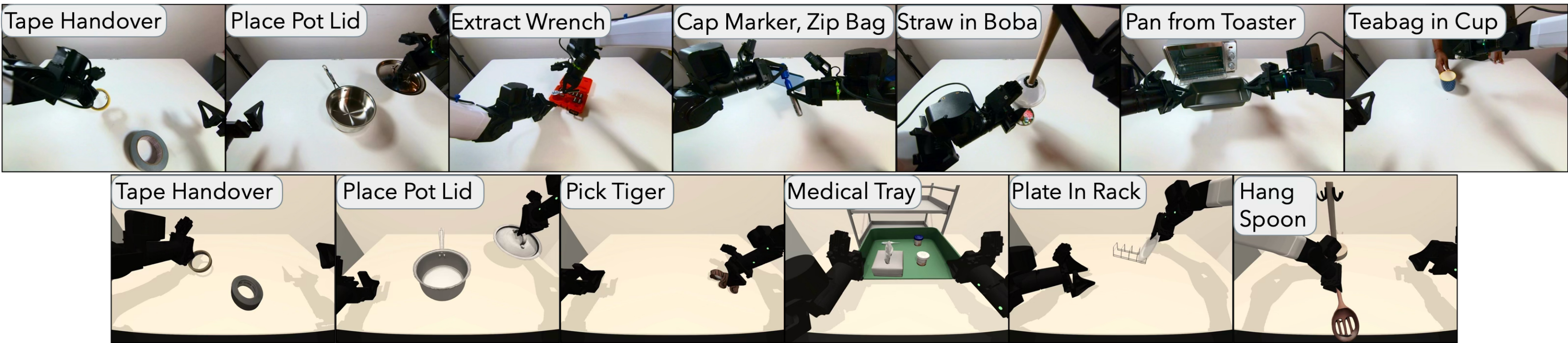}
    \caption{We collect teleop demonstrations for 7 real and 6 simulated tasks. Selected images show the raw perspective of the left stereo camera. Real-world tasks contain precise manipulation requirements, multi-stage behavior, articulated object interaction, and manipulation across the entire workspace. See the Appendix for a full breakdown of task stages and success criteria. }
    \label{fig:tasks}
    \vspace{-1.5em}
\end{figure*}

\section{Experiments}
\sysname{} is implemented using a bimanual, 14-DoF I2RT YAM manipulator with parallel-jaw grippers. We mount a static stereo camera between the arms 35\,cm above the base joint, angled 45$^\circ$ downwards for full observability of the table. Each eye captures a 120$^\circ$ diagonal FOV at $1600\times1200$ resolution. We use a custom leader arm modified from GELLO~\cite{wu2023gello} and TRLC~\cite{trlc2026} to teleop the robot.

\paragraph{Simulation Environment}
We build a simulation environment in MuJoCo~\cite{todorov2012mujoco} to conduct repeatable, large-volume experiments. Note that our objective is \textit{not} sim-to-real transfer, nor do we propose to use physics simulation to learn eye gaze, but rather we strive for a digital twin with which to conduct careful method comparisons using the same training pipeline as in real. We replicate the dimensions of the YAM manipulators, and use the public calibrated PD and mass-inertial constants in MJ-Playground~\cite{zakka2025mujocoplayground}. We implement non-pinhole camera rendering in MuJoCo by rendering pinhole images and warping them to exactly match the camera characteristics of each camera. We collect teleoperation with GELLOs through a custom VR interface which streams visuals to the operator in a Quest 3 headset, similar to prior VR teleop systems. All simulation code will be made public.

\paragraph{Tasks and Data}
In total we collected teleoperation data for \textbf{7} physical tasks and \textbf{6} simulated tasks as shown in Figure~\ref{fig:tasks}. We focus on 5 multi-stage real tasks which involve bimanual coordination (e.g. object handover, coordinated grasping), fine-grained tolerance requirements (e.g. marker capping, zipping, boba straw insertion), and long-horizon behavior (e.g. opening a toaster oven, taking out a pan, pushing the shelf back in, closing the door). Due to the challenges of simulating contact-rich physical interactions in real-time, our simulated tasks aim to evaluate a range of basic pick and place behavior. We collect between 10 and 53 minutes of data per task. Full task definitions, stage visualizations, train distribution visualizations, and dataset size details are provided in the Appendix.

\section{Results}
We evaluate how well \sysname{} performs compared to SOTA baselines. To construct carefully controlled comparisons, we choose to collect all data ourselves and train all policies from scratch on the \emph{exact same} demonstrations. This avoids obfuscation from unknown pretraining distributions of off-the-shelf models, reduces compute demand, and increases the tractable volume of experiments by an order of magnitude. 

In \textbf{real}, we train every policy with 3 seeds and randomly select one for evaluation. For each task, we evaluate all policies on 25 scene configurations selected by placing objects within the training data region, but not exactly matching train datapoints. During evaluation, we overlay the live camera stream on the test positions and manually align objects so all policies share the \textit{exact same} test positions. See the Appendix for a visualization of these test distributions. Across all real-world physical trials of baselines and ablations, we evaluate over 1000 policy rollouts.

In \textbf{sim}, every baseline and \sysname{} policy is trained across 3 seeds, of which we report the maximum performance to prune failed seeds. See Appendix for full statistics. All sim evaluations are run on the same 100 random configurations, all unseen during training but drawn from the same distribution. Each policy is trained from scratch on single-task data, and executed using the same temporal ensembling weights and inference code. The gaze policy is given 1 second to allow gaze to settle before acting. Per-stage success is programmatically judged using manually specified MuJoCo physics requirements, and each task has a failure timeout based on the average length of demonstrations.

\paragraph{Baselines}
We collect all our data with stereo ego and wrist cameras, meaning \sysname{} and baselines share the same training data. \textit{No Gaze} takes in the raw stereo input stream, which has the exact same observability but without fixation, and \textit{Ego + Wrist} mimics the prevailing bimanual setup using two wrist cameras and the left image from the stereo pair. 
All baselines are trained with ACT~\cite{zhao2023learning} and share the same architecture and training hyperparameters as \sysname{}. We augment baseline input images with 95\% center-crop, and 10\% translation and rotation randomization to make them less sensitive to placement error, as in~\cite{black2410pi0}. To select the best action space, we test baselines with end-effector (EE) relative SE(3), world-absolute SE(3), joint chunk-relative, and joint chunk-absolute. Since EE-relative SE(3) performs the best in sim (see Appendix), we use that for real evaluations. Baselines are trained with a chunk size of 1 second as suggested in ACT~\cite{zhao2023learning} and \sysname{} is trained with 1.5 seconds. We found a slight performance gain from 1.5s chunks with fixation but no such gain for baselines when swept in simulated tasks.

\begin{figure}[t]
    \centering
    \includegraphics[width=\linewidth]{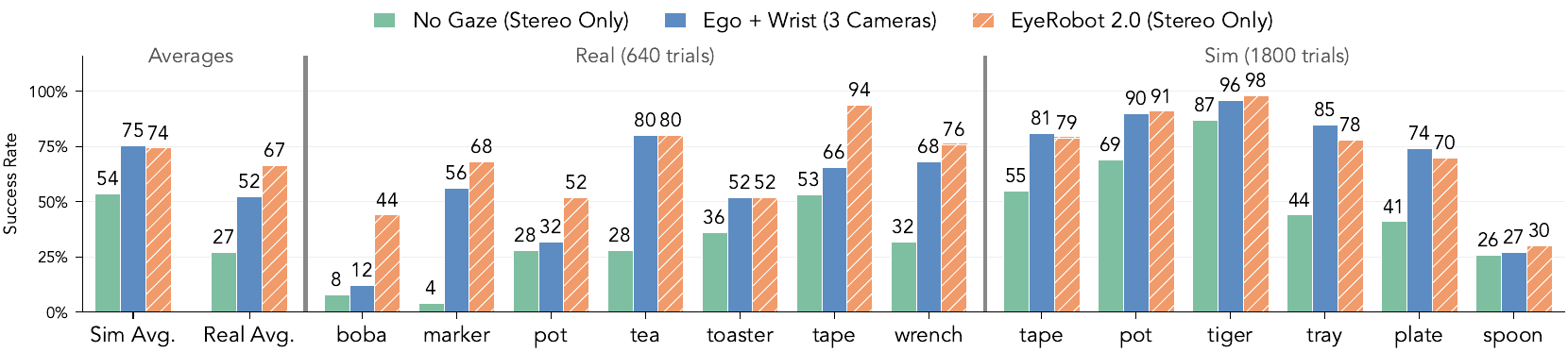}
    \caption{\textbf{End-to-end success on real and sim tasks.} Each real task is evaluated on 25 trials for each policy, all sim tasks over 100. Full per-stage conditional success is reported in the Appendix.}
    \label{fig:results}
\end{figure}

\paragraph{\textit{How does \sysname{} compare to gaze-free stereo?}} This baseline corresponds to training standard policies with static ego-only stereo inputs without wrist cameras.
\sysname{} outperforms static stereo policies significantly across all 13 tasks, on average by $+20\%$ in sim and $+40\%$ in real (Fig.~\ref{fig:results}). Ego-only policies have a strong tendency to overfit to the training positions, an effect which can particularly be seen in the \texttt{tape\_handover} task, as data was collected in a regular $8\times8$ grid, or the \texttt{tea} task which requires servoing around the entire workspace. In addition, ego policies lack the visual acuity to competently accomplish precise tasks like marker insertion (4\%) and straw insertion (8\%). Despite having access to the exact same stereo stream, by foveating on relevant objects \sysname{} achieves 68\% and 44\% end-to-end success on these two tasks. The failure modes of static stereo policies tend to cluster around task stages that require precision (see Sankey diagrams in Appendix), as well as significantly more overfitting behavior in tasks with wide spatial distribution (tape handover).

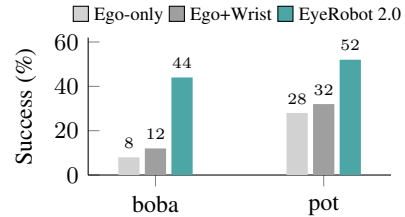
\begin{wrapfigure}{r}{0.4\linewidth}
    \vspace{-1.2em}
    \centering
    \begin{tikzpicture}
    \begin{axis}[
        ybar=2pt, bar width=8pt, width=\linewidth, height=3.4cm,
        ymin=0, ymax=62, ytick={0,20,40,60},
        ylabel={Success (\%)}, ylabel style={font=\footnotesize},
        symbolic x coords={boba,pot}, xtick=data,
        tick label style={font=\footnotesize},
        legend style={font=\scriptsize, at={(0.5,1.02)}, anchor=south, legend columns=3, draw=none},
        legend image code/.code={\draw[#1] (0cm,-0.08cm) rectangle (0.2cm,0.12cm);},
        nodes near coords, nodes near coords style={font=\tiny},
        enlarge x limits=0.4, axis lines*=left,
    ]
    \addplot[draw=none, fill=gray!35] coordinates {(boba,8) (pot,28)};
    \addplot[draw=none, fill=gray!75] coordinates {(boba,12) (pot,32)};
    \addplot[draw=none, fill=teal!70] coordinates {(boba,44) (pot,52)};
    \legend{Ego-only, Ego+Wrist, \sysname{}}
    \end{axis}
    \end{tikzpicture}
    \vspace{-0.6em}
    \caption{\textbf{Wrist occlusion.} When grasped objects block the wrist cameras, ego+wrist falls to near the stereo baseline while \sysname{} maintains performance.}
    \label{fig:occlusion}
    \vspace{-1.2em}
\end{wrapfigure}
\paragraph{\textit{How does \sysname{} compare against SOTA ego + wrist policies?}} 
On precise physical tasks where the wrist camera maintains full visibility (\texttt{marker}, \texttt{tea}, \texttt{toaster}, \texttt{wrench}), both perform similarly (Fig.~\ref{fig:results}).
In tasks where grasped objects occlude the wrist cameras (\texttt{boba}, \texttt{pot}), ego+wrist performance collapses to near the gaze-free stereo baseline, while \sysname{} maintains strong performance, overall a $2\times$ improvement (Fig.~\ref{fig:occlusion}). In the \texttt{tape\_handover} task, wrist camera policies tend to overfit to the grid training distribution similarly to ego-only policies because the wrist camera on the gripper holding the yellow tape is occluded until just before placement, by which point the tape is already in an unrecoverable position.

\paragraph{\textit{Does fixation improve robustness to visual distractors?}} We evaluate all policies in real on \texttt{tea}, \texttt{wrench}, and \texttt{tape} with colorful distractor objects randomly scattered on the table and report their end-to-end success along with first stage success in Table~\ref{tab:distractors}. All comparisons use the exact same object positions to ensure comparability. The wrist camera policy often reaches for the largest or most distinct object in its view. Once it begins reaching, the policy cannot recover since the wrist camera loses sight of the target. In contrast, \sysname{} reaches for the correct object when its gaze is correctly directed at it, and mostly misses grasps when it is looking in the wrong place. This shows most clearly in first-grasp success, which is significantly higher with active gaze. Decoupling vision and action could improve robustness in cluttered scenes by literally ignoring irrelevant information.

\begin{table}[t]
    \centering
    \small
    \begin{tabular}{l cc cc cc}
    \toprule
    & \multicolumn{2}{c}{wrench} & \multicolumn{2}{c}{tea} & \multicolumn{2}{c}{tape} \\
    \cmidrule(lr){2-3} \cmidrule(lr){4-5} \cmidrule(lr){6-7}
    \textbf{Policy} & 1st grasp & success & 1st grasp & success & 1st grasp & success \\
    \midrule
    Ego-only    & 3/25  & 0/25 & 3/25  & 0/25 & 6/25  & 0/25 \\
    Ego + Wrist & 10/25 & 4/25 & 11/25 & \textbf{4/25} & 14/25 & 0/25 \\
    \textbf{\sysname{}} & \textbf{20/25} & \textbf{5/25} & \textbf{13/25} & \textbf{4/25} & \textbf{23/25} & \textbf{8/25} \\
    \bottomrule
    \end{tabular}
    \vspace{.2em}
    \caption{\textbf{Real-world performance with distractors.} \sysname{} is better able to ignore distractor objects because of its physical attention directed towards relevant objects.}
    \label{tab:distractors}
\end{table}

\paragraph{Gaze analysis} See Fig.~\ref{fig:probabilities} for a visualization of target selector probabilities over the full course of the marker task. \sysname{} first fixates on the marker cap, then switches to the marker just before picking it up and keeps its gaze there throughout capping. Overall, it learns intuitive gaze switching, for example looking at an object until just after it has been grasped. Interestingly, it sometimes uses static objects as a sort of anchor during manipulation, for example during spans of the marker task it gazes at the pencil pouch which during data collection was always in a roughly straight-ahead position.

     \begin{table}[t]
    \centering
    \small
    \begin{tabular}{lccccccc|c}
    \toprule
    \textbf{Policy} & boba & marker & pot & tea & toaster & tape & wrench & \textbf{Mean} \\
    \midrule
    \sysname{} $-$foveation & \textbf{12/25} & 5/25 & 10/25 & 12/25 & 9/25 & 47/64 & 15/25 & 46.5\% \\
    \sysname{} $-$stereo & 10/25 & 8/25 & 10/25 & 15/25 & 6/25 & 51/64 & 16/25 & 48.5\% \\
    \textbf{\sysname{}} & 11/25 & \textbf{17/25} & \textbf{13/25} & \textbf{20/25} & \textbf{13/25} & \textbf{60/64} & \textbf{19/25} & \textbf{66.5\%} \\
    \bottomrule
    \end{tabular}
    \vspace{.2em}
    \caption{\textbf{Real-world ablations} show that foveation and stereo inputs are both important.}
    \label{tab:ablations}
    \vspace{-2em}
\end{table}

\subsection{Ablations}

\paragraph{No foveated image architecture} We ablate the multi-scale image pyramid and train both gaze and gripper policies on one peripheral view for each eye (Tab.~\ref{tab:ablations}). Overall this harms performance by 20\%, with the largest gap on precise physical tasks which require detailed image understanding.  Interestingly, in simulation this ablation performs the same as \sysname{}. This may be because robot execution in simulation has little to no variance, whereas real-world execution varies from trial to trial and so demands more closed-loop visual servoing.

\paragraph{Fixation-centric actions}
We test an ablation of \sysname{} in simulation which uses SE(3) actions and proprioception in the \textit{world} frame rather than a fixation-relative frame. This ablation performs on par with the gaze-free ego baseline ($-21\%$). Including gaze while still allowing a global output space for the policy seems to still afford overfitting to specific trajectory positions in this low-data regime.

\paragraph{Stereo vs.\ Mono}
We train an ablation which only has one input image and no input fixation depth. This ablation performs on average $18\%$ worse than \sysname{} across all real-world tasks, indicating that the model benefits from access to binocular information.

\section{Limitations and Future Work}
\label{sec:limitations}
\sysname{} currently only trains task-specific gaze and gripper policies. An interesting direction for future extension is to investigate how to adapt the framework to maintain multi-task performance. The usage of a target selector means that gaze would need to be trained on a sufficiently large corpus to allow generic prompts, which may necessitate different architectures for generating prompts such as a vision-language model. Separately, in this work we do not control head or neck movement, and only propose how to coordinate fine-grained eye movements provided a good head position is already available. Finally, while object-level pretraining is sufficient for many manipulation tasks, it cannot yet train sub-object-level gaze which may provide more benefits for manipulation.

\section{Conclusion}
\label{sec:conclusion}
This work presents \sysname{}, a robot learning framework for fine-grained bimanual manipulation using only a fixed stereo camera that leverages active visual fixation to direct attention towards task-relevant features. Our experiments suggest that active fixation more effectively leverages stereo data than non-active stereo baselines, strongly outperforming them across a range of high-precision physical tasks and matching or exceeding the performance of wrist cameras. Ablating any of its components results in a performance drop, indicating it utilizes multi-resolution image inputs, fixation-centric actions, and stereo information to effectively facilitate fine-grained real-world manipulation. This suggests exciting potential for systems which use active visual fixation for fine-grained manipulation instead of wrist cameras.

\clearpage

\begin{center}
    {\LARGE\bf Appendix}
\end{center}

\appendix
\section{Additional Results}

\begin{figure*}[t]
    \centering
    \begin{tabular}{ccccccc}
        \rotatebox{-90}{\includegraphics[width=0.25\textheight]{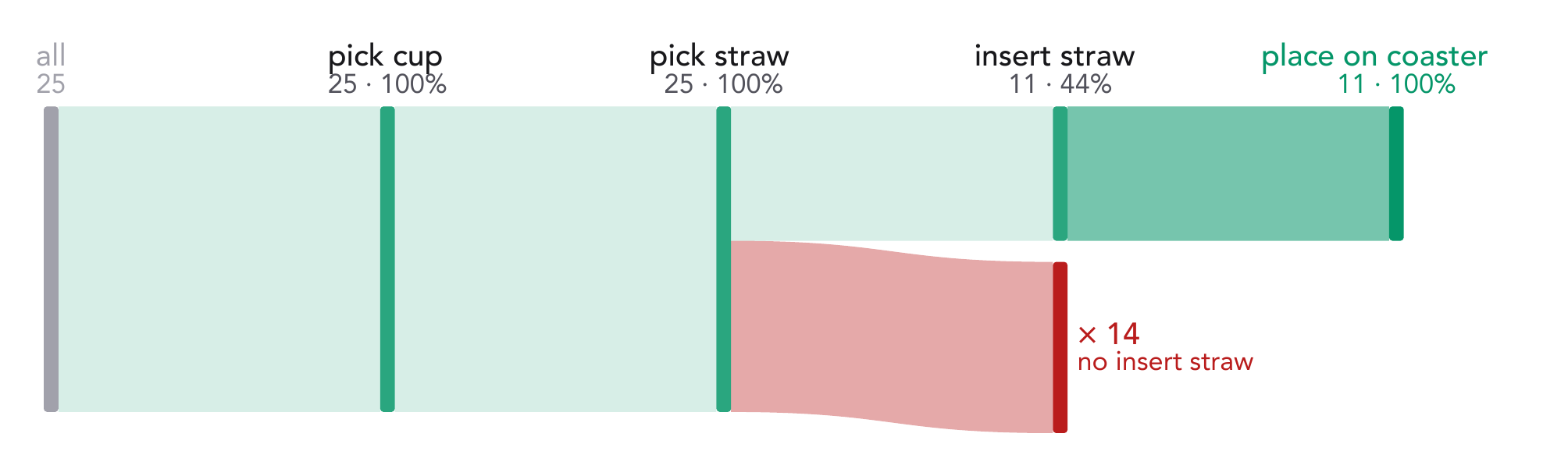}} &
        \rotatebox{-90}{\includegraphics[width=0.25\textheight]{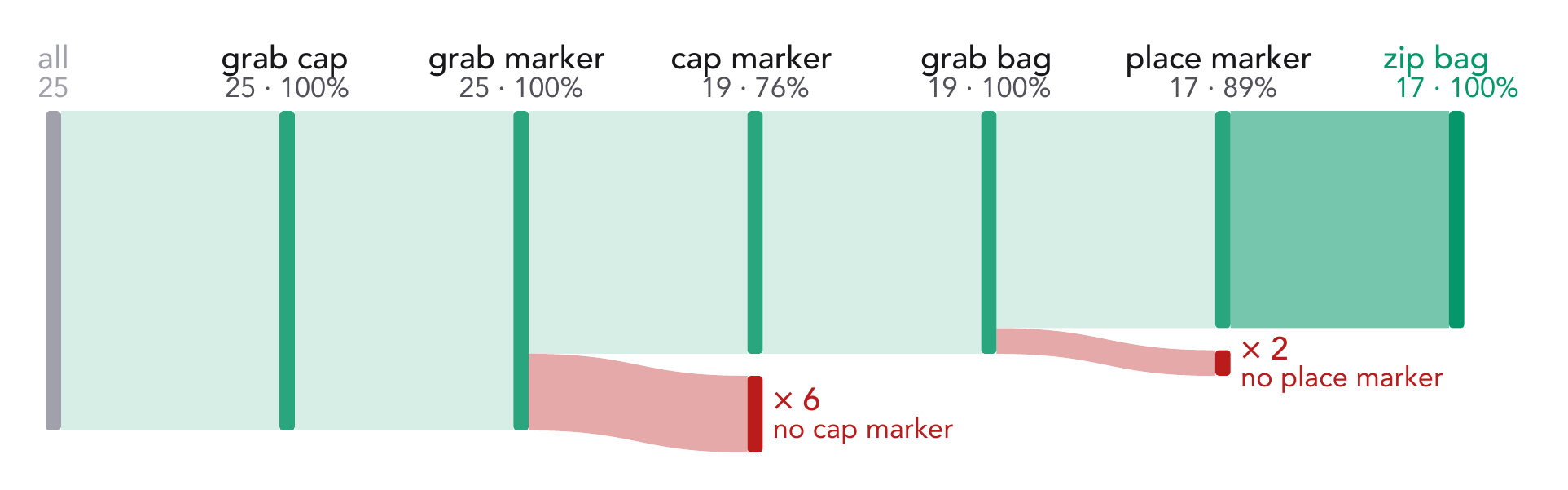}} &
        \rotatebox{-90}{\includegraphics[width=0.25\textheight]{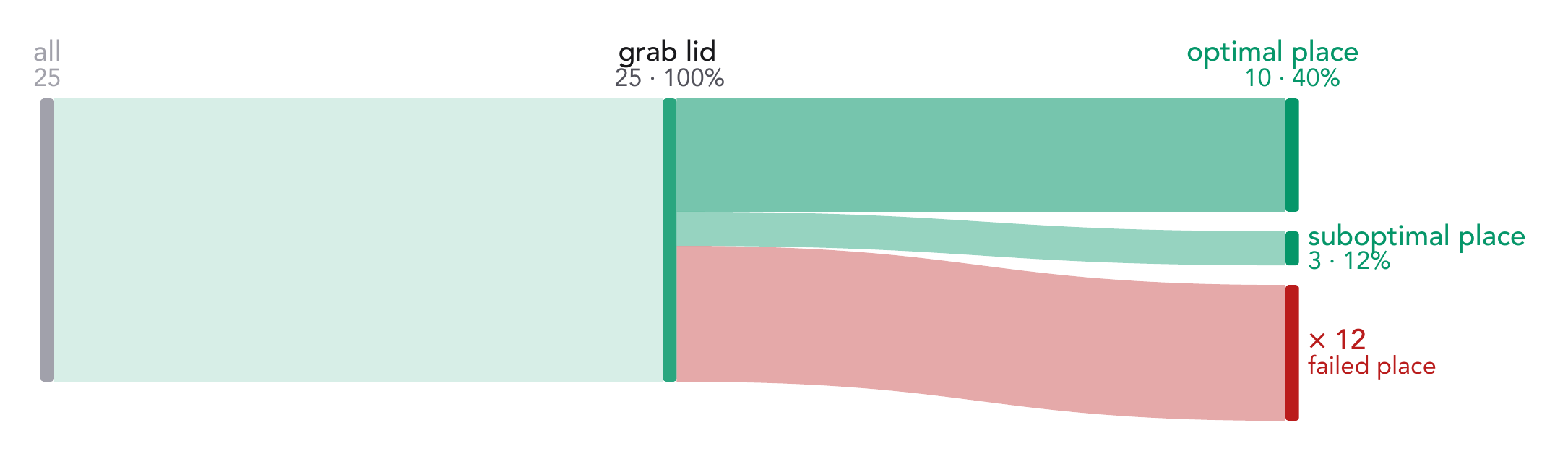}} &
        \rotatebox{-90}{\includegraphics[width=0.25\textheight]{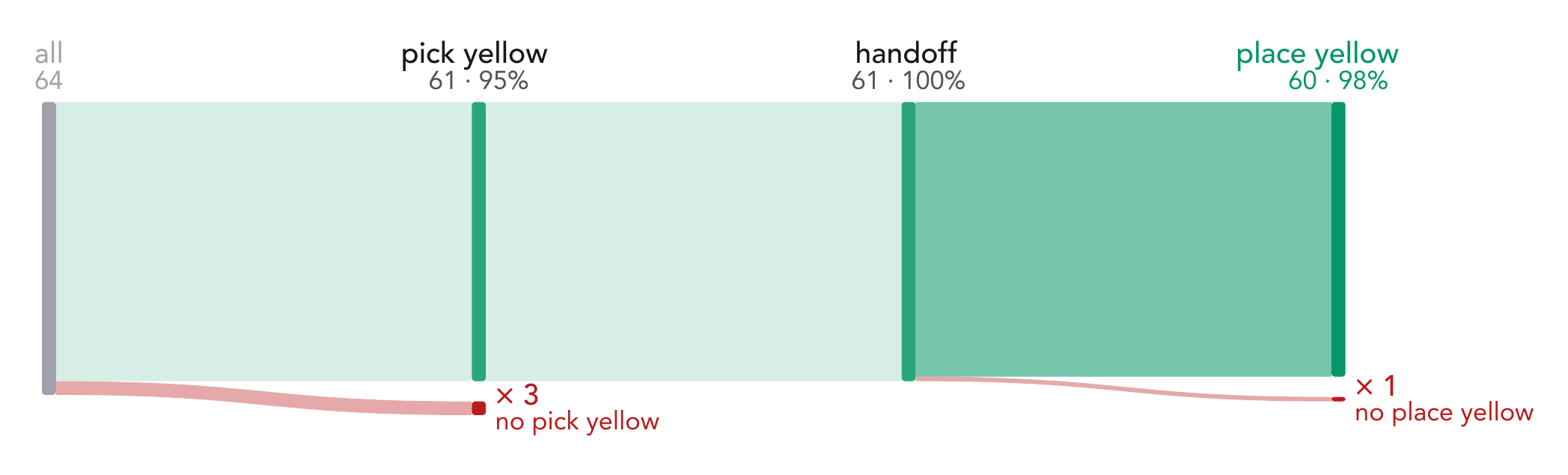}} &
        \rotatebox{-90}{\includegraphics[width=0.25\textheight]{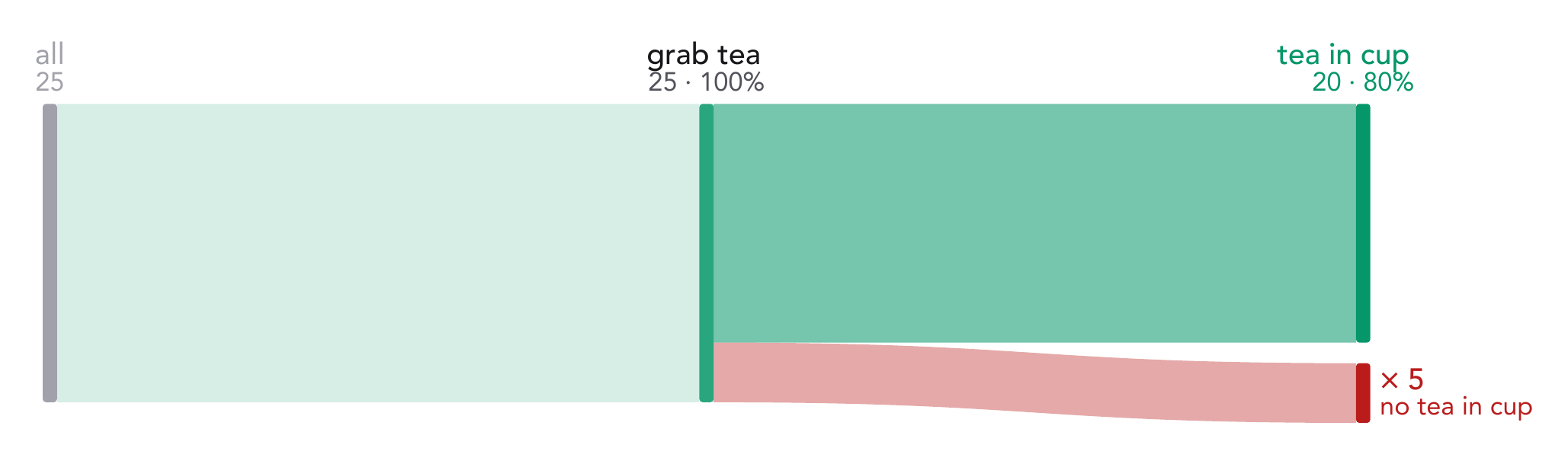}} &
        \rotatebox{-90}{\includegraphics[width=0.25\textheight]{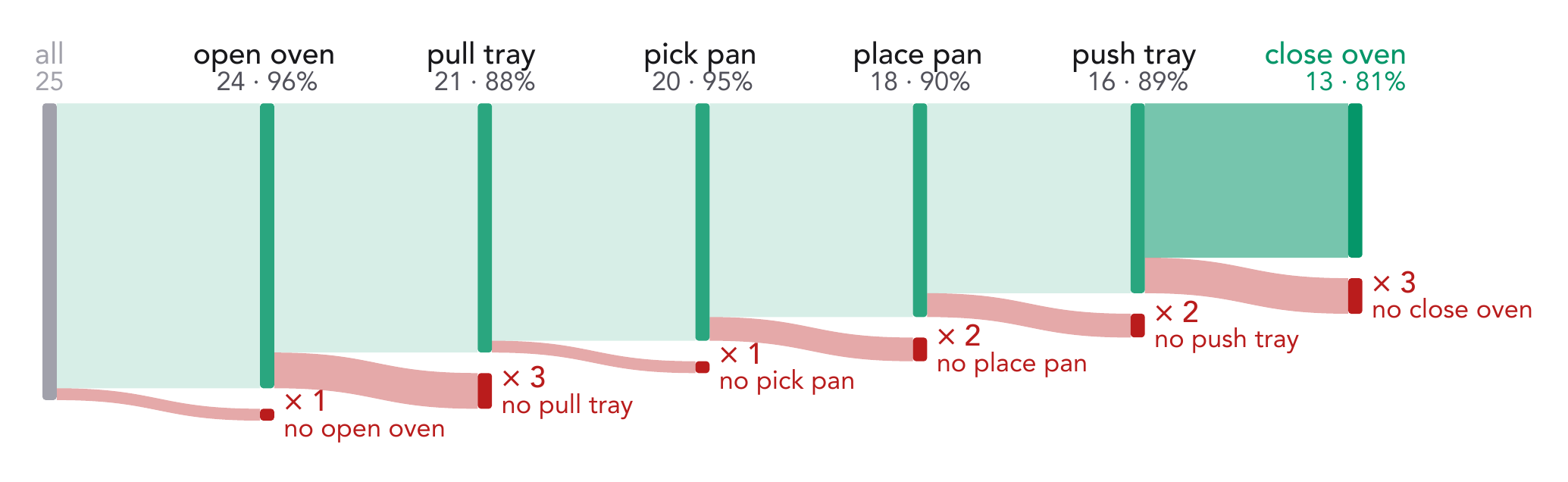}} &
        \rotatebox{-90}{\includegraphics[width=0.25\textheight]{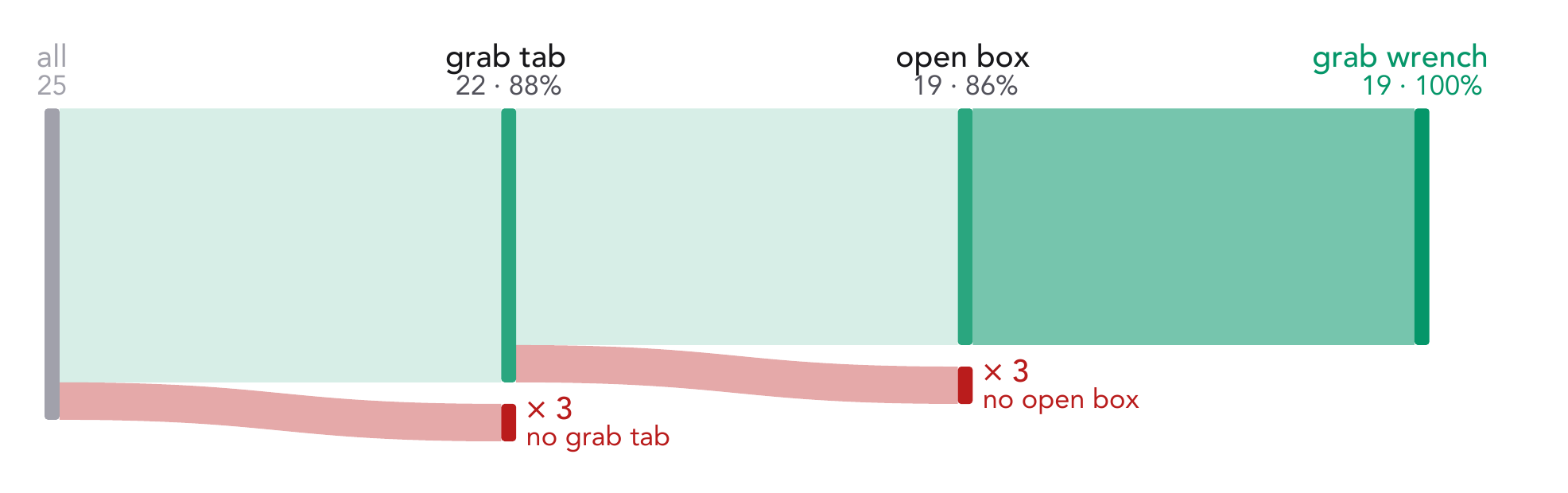}} \\
        Boba & Marker & Pot Place & Tape & Tea & Toaster & Wrench
    \end{tabular}
    \caption{\textbf{Per-stage conditional success for \sysname{}} across 25 physical trials per task.}
    \label{fig:success_funnels_avf}
\end{figure*}

\begin{figure*}[t]
    \centering
    \begin{tabular}{ccccccc}
        \rotatebox{-90}{\includegraphics[width=0.25\textheight]{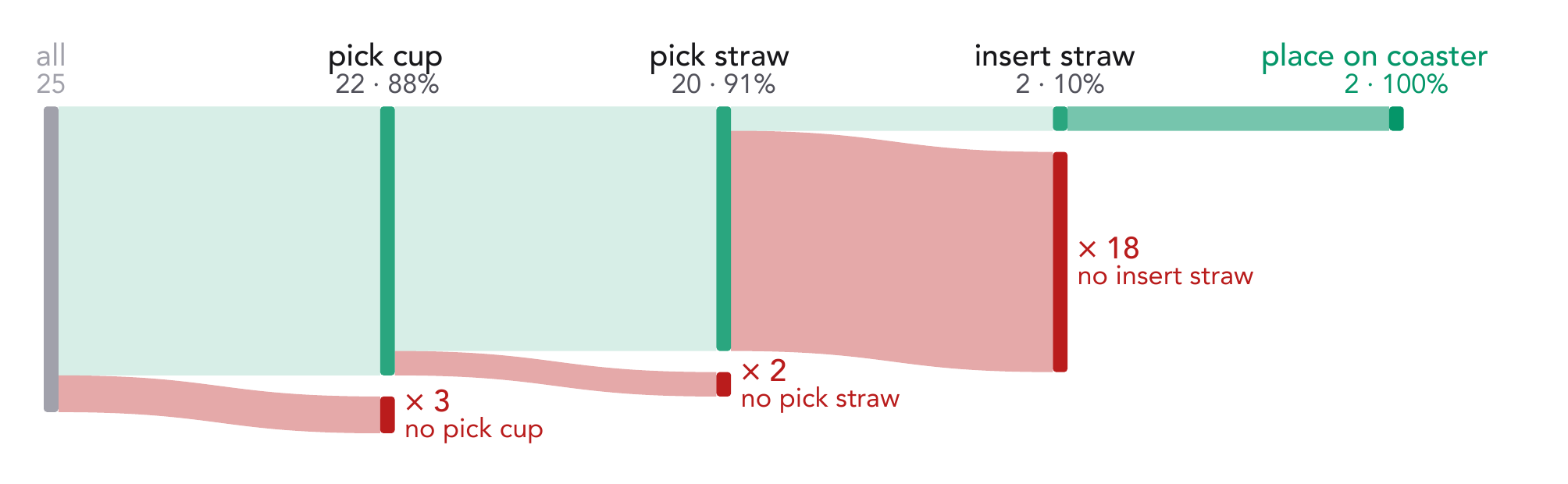}} &
        \rotatebox{-90}{\includegraphics[width=0.25\textheight]{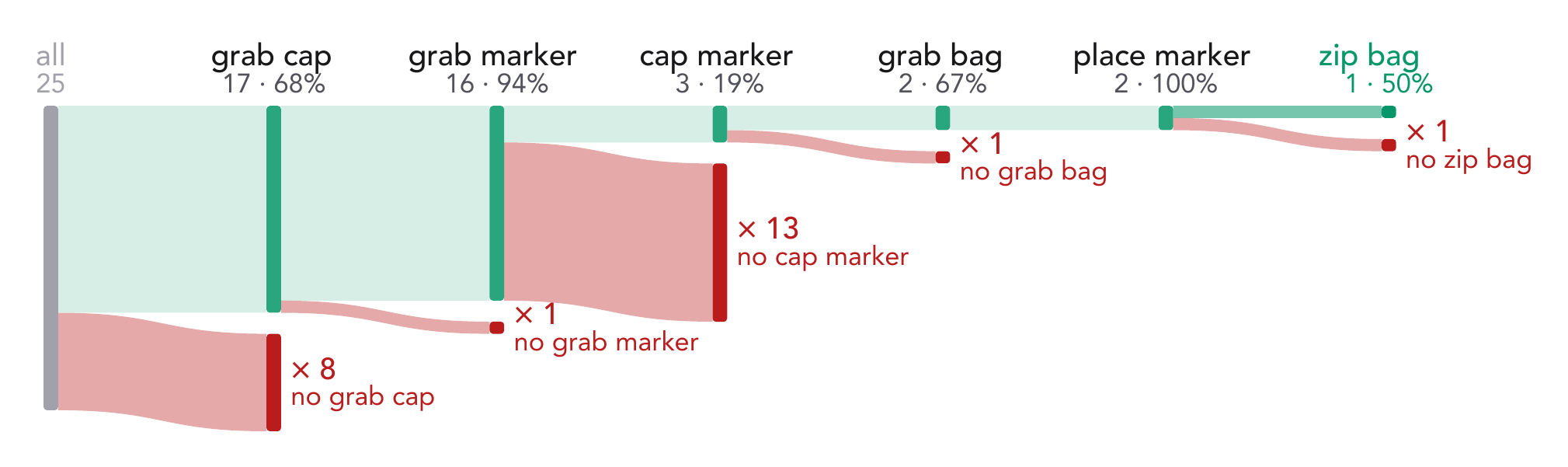}} &
        \rotatebox{-90}{\includegraphics[width=0.25\textheight]{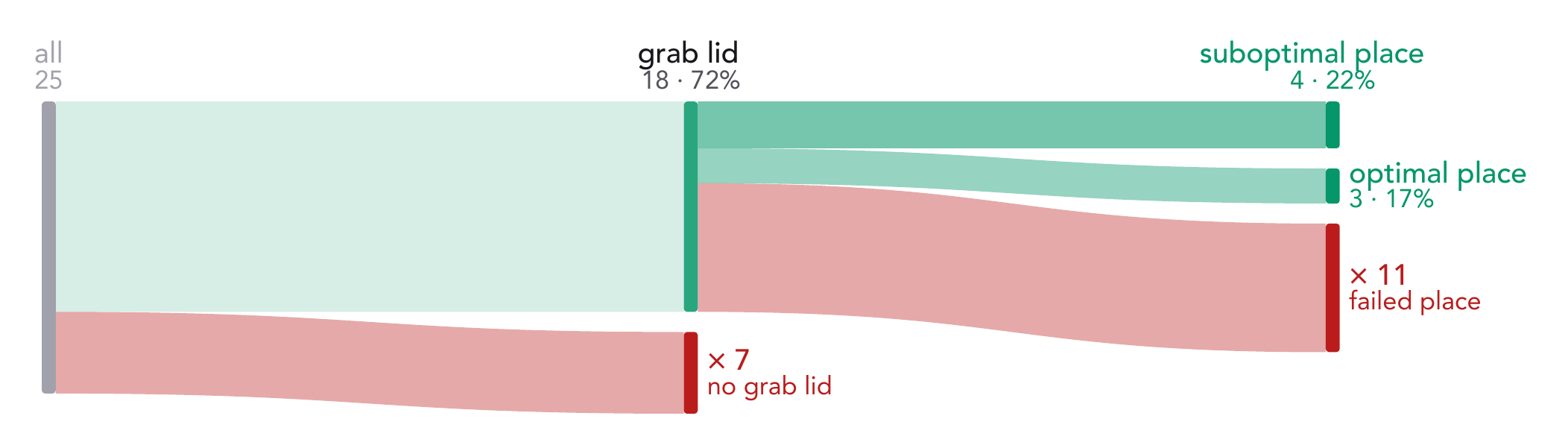}} &
        \rotatebox{-90}{\includegraphics[width=0.25\textheight]{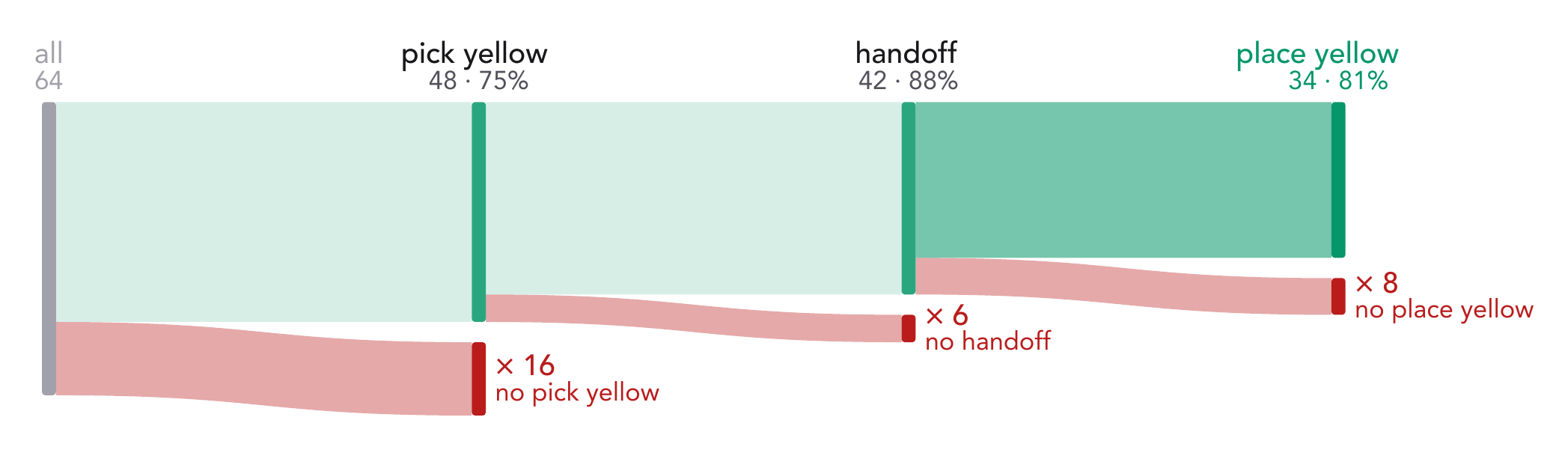}} &
        \rotatebox{-90}{\includegraphics[width=0.25\textheight]{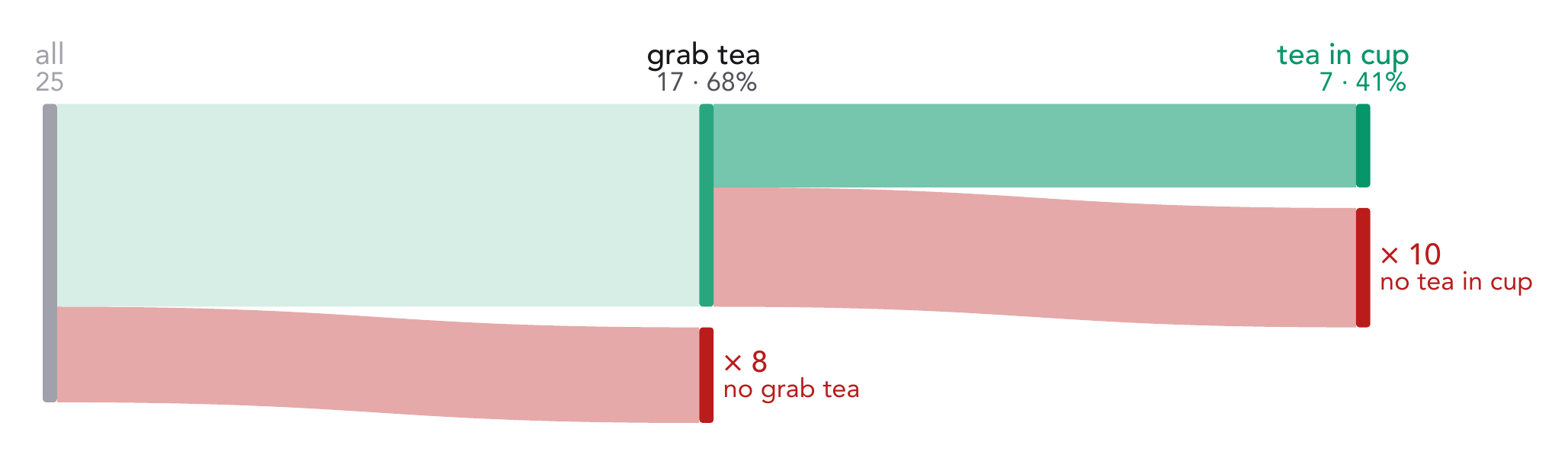}} &
        \rotatebox{-90}{\includegraphics[width=0.25\textheight]{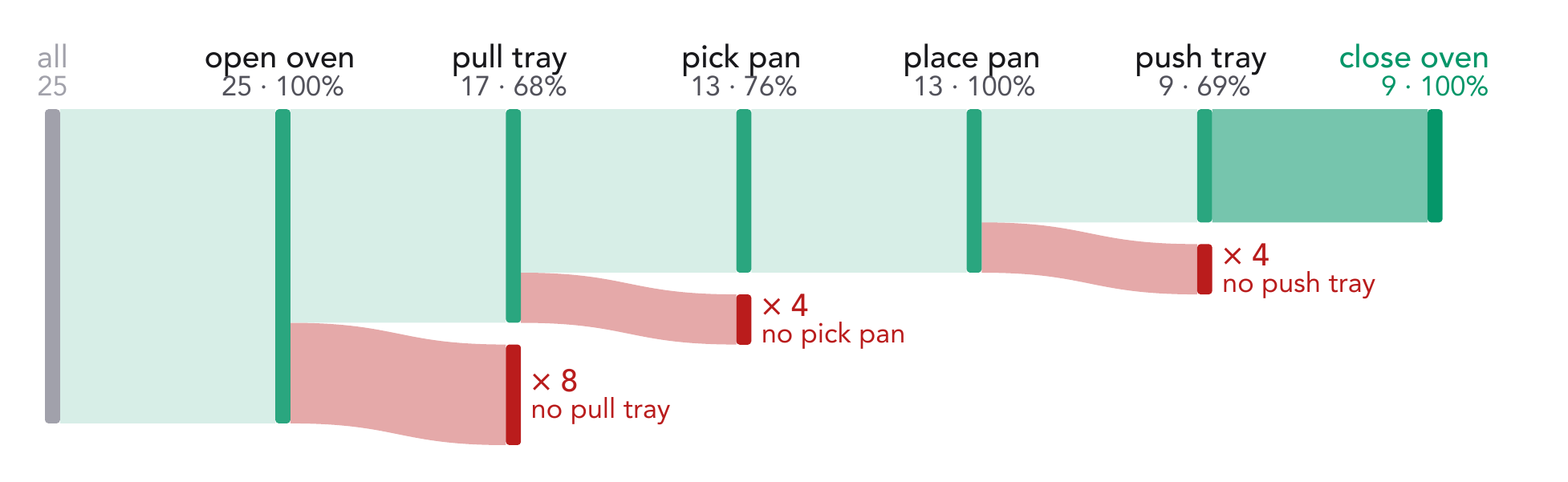}} &
        \rotatebox{-90}{\includegraphics[width=0.25\textheight]{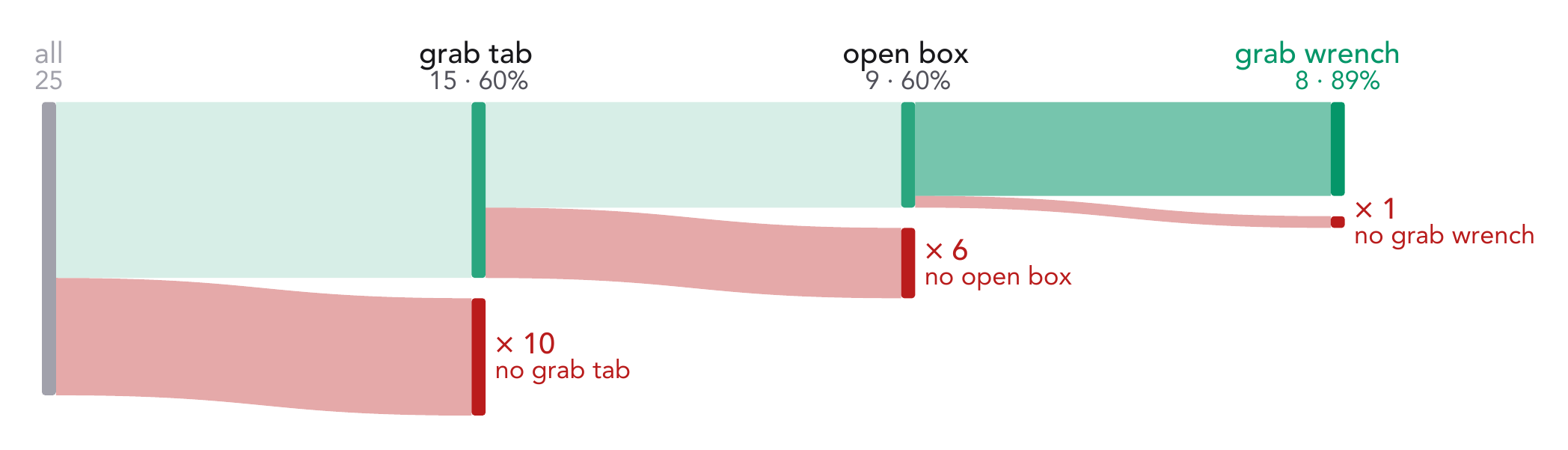}} \\
        Boba & Marker & Pot Place & Tape & Tea & Toaster & Wrench
    \end{tabular}
    \caption{\textbf{Per-stage conditional success for Ego-only gaze-free baseline} across 25 physical trials per task.}
    \label{fig:success_funnels_ego}
\end{figure*}

\begin{figure*}[t]
    \centering
    \begin{tabular}{ccccccc}
        \rotatebox{-90}{\includegraphics[width=0.25\textheight]{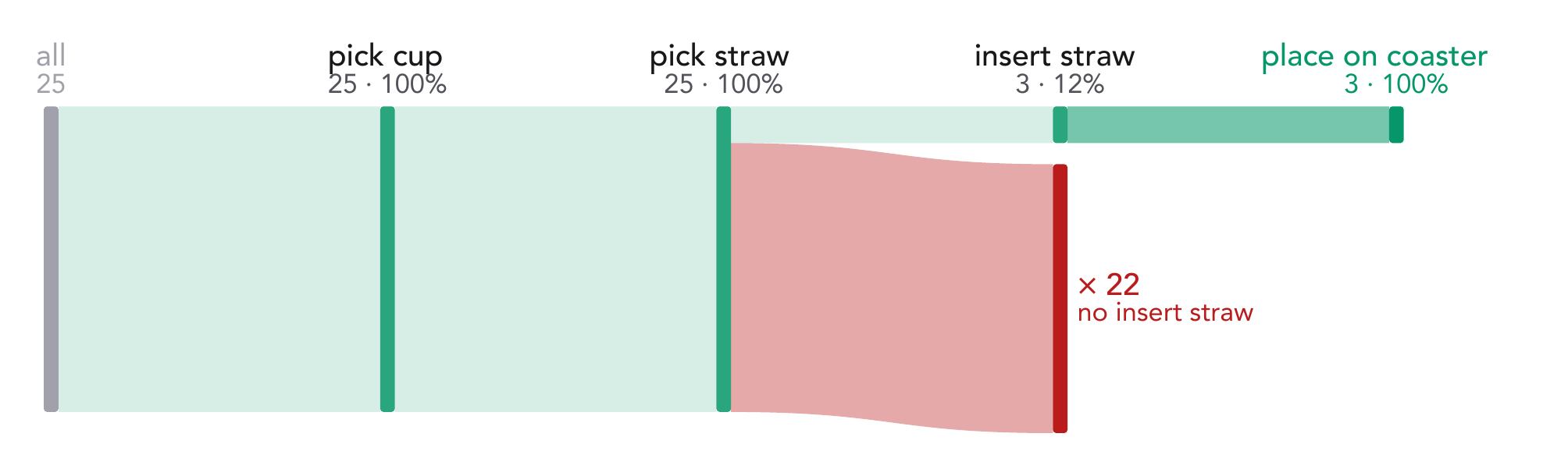}} &
        \rotatebox{-90}{\includegraphics[width=0.25\textheight]{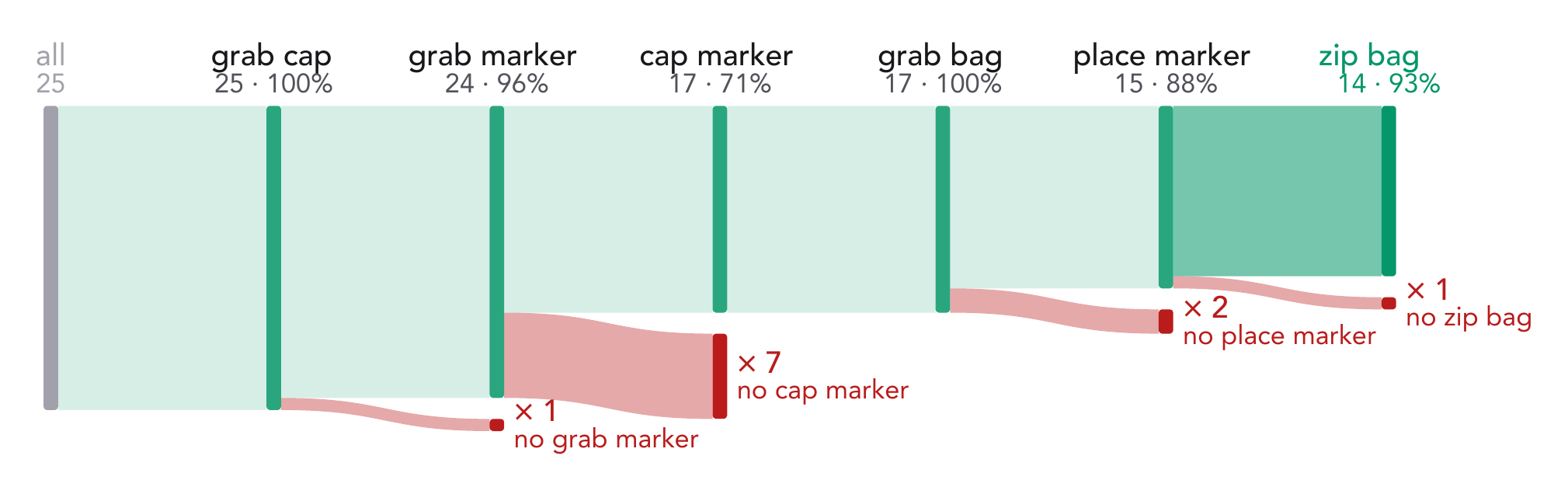}} &
        \rotatebox{-90}{\includegraphics[width=0.25\textheight]{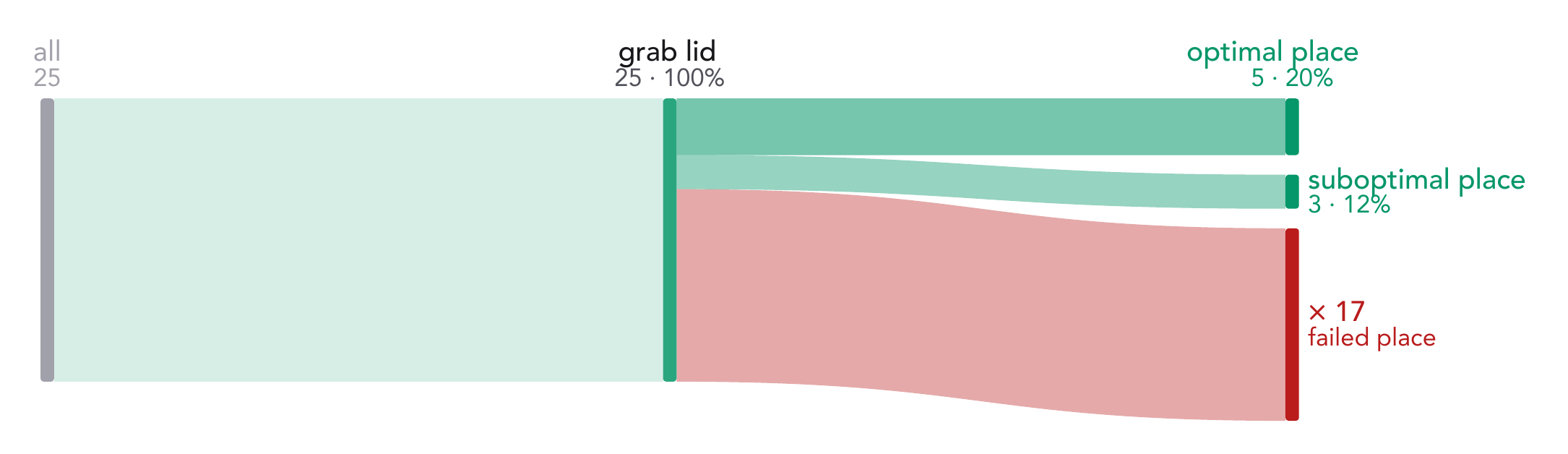}} &
        \rotatebox{-90}{\includegraphics[width=0.25\textheight]{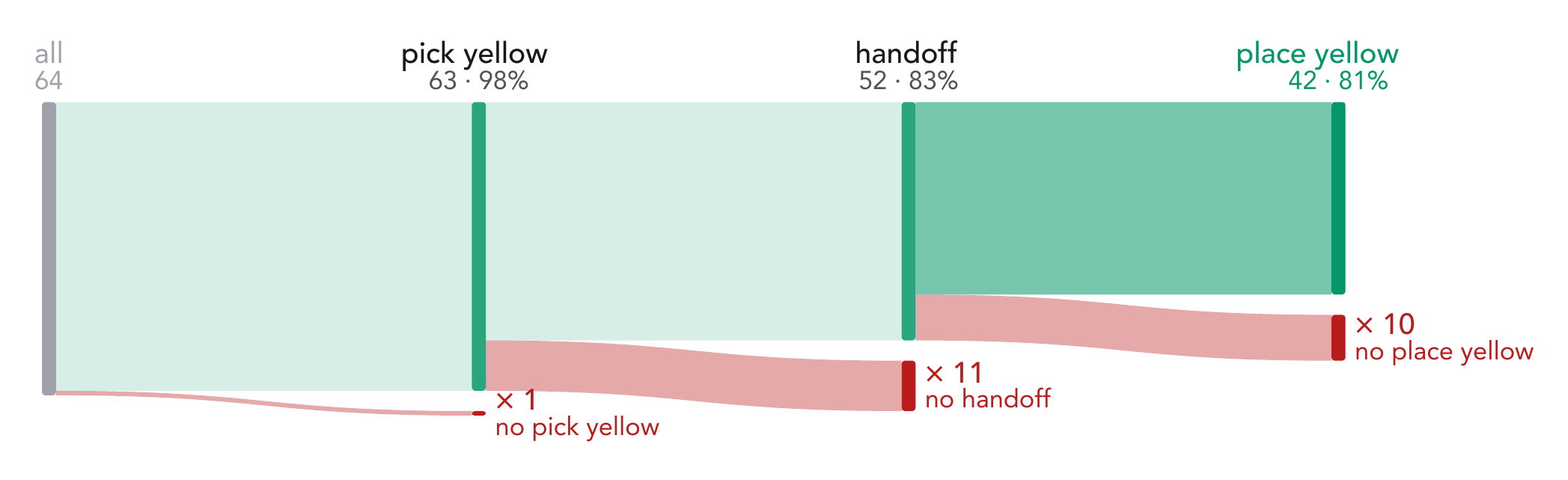}} &
        \rotatebox{-90}{\includegraphics[width=0.25\textheight]{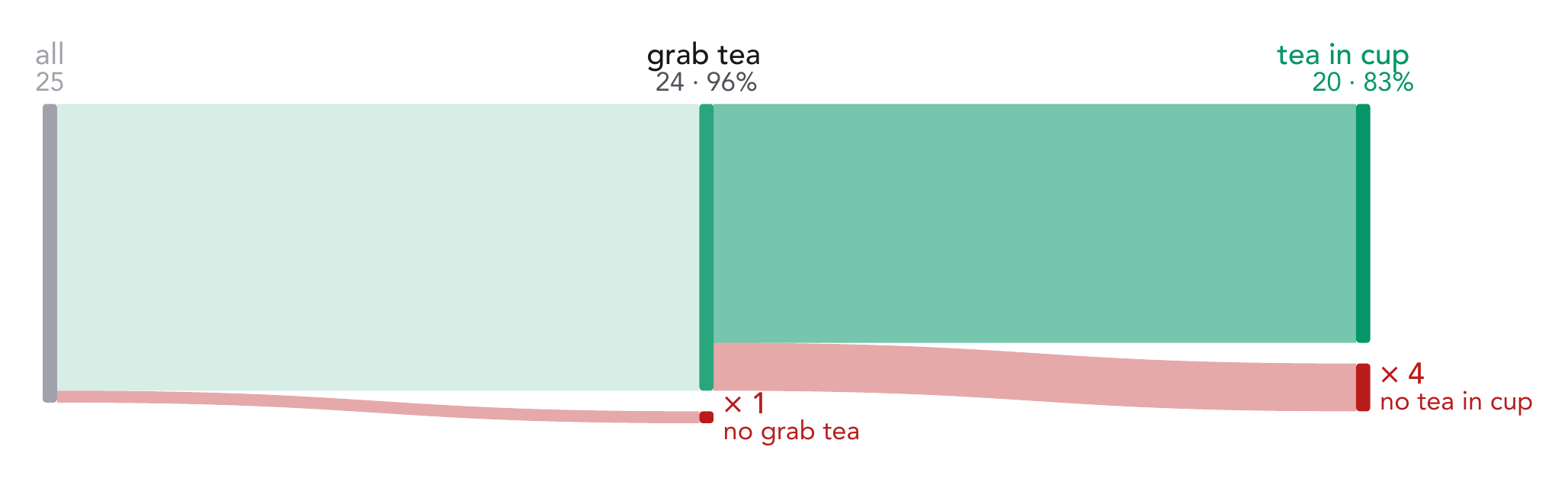}} &
        \rotatebox{-90}{\includegraphics[width=0.25\textheight]{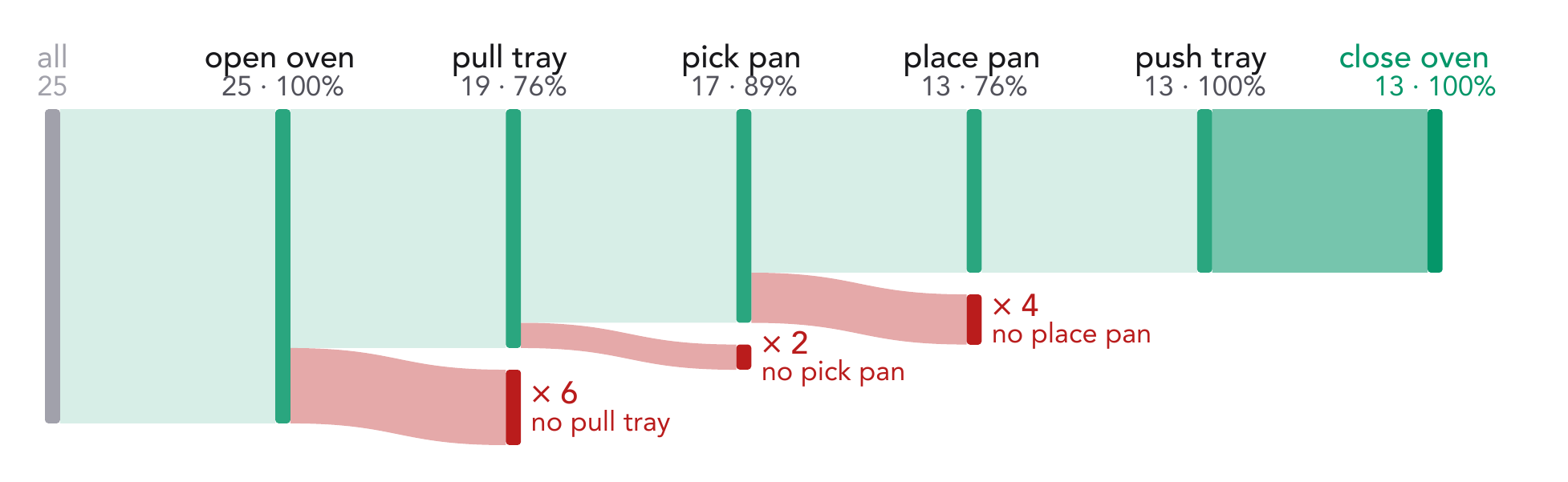}} &
        \rotatebox{-90}{\includegraphics[width=0.25\textheight]{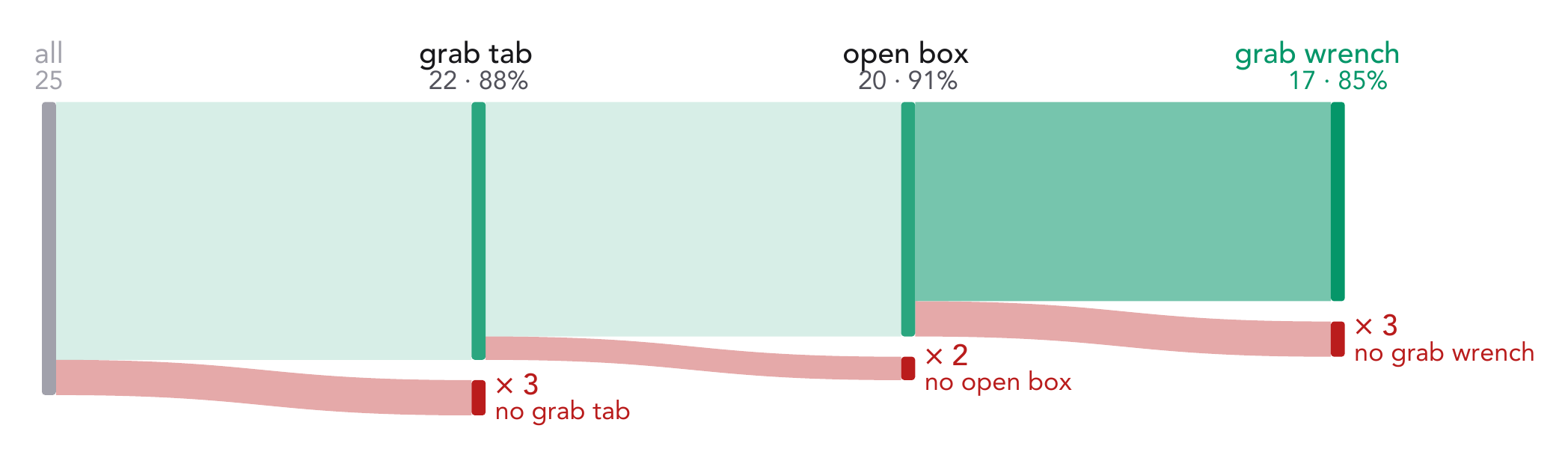}} \\
        Boba & Marker & Pot Place & Tape & Tea & Toaster & Wrench
    \end{tabular}
    \caption{\textbf{Per-stage conditional success for Ego$+$Wrist baseline} across 25 physical trials per task.}
    \label{fig:success_funnels_egowrist}
\end{figure*}

\paragraph{\textit{Failure modes analysis}} Please see the included Sankey diagrams~(Figs.~\ref{fig:success_funnels_avf},~\ref{fig:success_funnels_ego},~\ref{fig:success_funnels_egowrist}) for a visualization of per-stage success breakdown for \methodname{} and baselines across all physical trials. The primary failure cases align with the most difficult, precision-requiring stages of each task, for example inserting the boba straw or capping the marker. In comparison, the ego-only baseline struggles more with these precise stages, and also has considerably more failures in ``easy'' portions of each task, for example picking up objects initially. With only a fixed input view the policy struggles to achieve the precision required to succeed, and even struggles more at basic picking, for example in the tea task only grasping the small teabag in 32\% of trials, while \methodname{} always grasped the teabag. The ego$+$wrist baseline recovers this grasping precision, but fails severely on task stages (pot on lid, boba insertion) where wrists become occluded by the grasped object. Even in tasks with unoccluded views from the wrist camera, wrist cameras have similar conditional success rates to \methodname{}, indicating that failures stem from a common modeling or data insufficiency which wrist cameras provide no advantage for.

\paragraph{\textit{How well do policy rankings in sim and real correlate?}} To measure sim-real correlation we collect 2 similar tasks (\texttt{pot} and \texttt{tape}) in sim and in real. While the absolute success rate in sim is not predictive of real-world success, policy rankings are maintained on these tasks between all baselines, and results indicate that sim may favor wrist cameras over physical trials, perhaps because EE-relative chunks are more sensitive to compounding action errors with imperfect real physics than simulated. Pot real-world success is notably lower overall because simulation has more forgiving physics for pot lid closure.

\paragraph{\textit{What fixation patterns does \methodname{} learn?}} Probabilities while inserting the marker into the bag are more ambiguous as all objects are in the same location. Quantitatively, we analyze per-stage gaze in simulation and report full task results in the Appendix. During the first task stage, 100\% of policies learned to look at the first object the robot grasps. In \texttt{tape\_handover}, \texttt{dish\_rack}, and \texttt{tray} the gaze switches to the second object, while in the \texttt{pot} and \texttt{spoon} tasks it remains on the first object.

\begin{figure}
    \centering
    \includegraphics[width=0.8\linewidth]{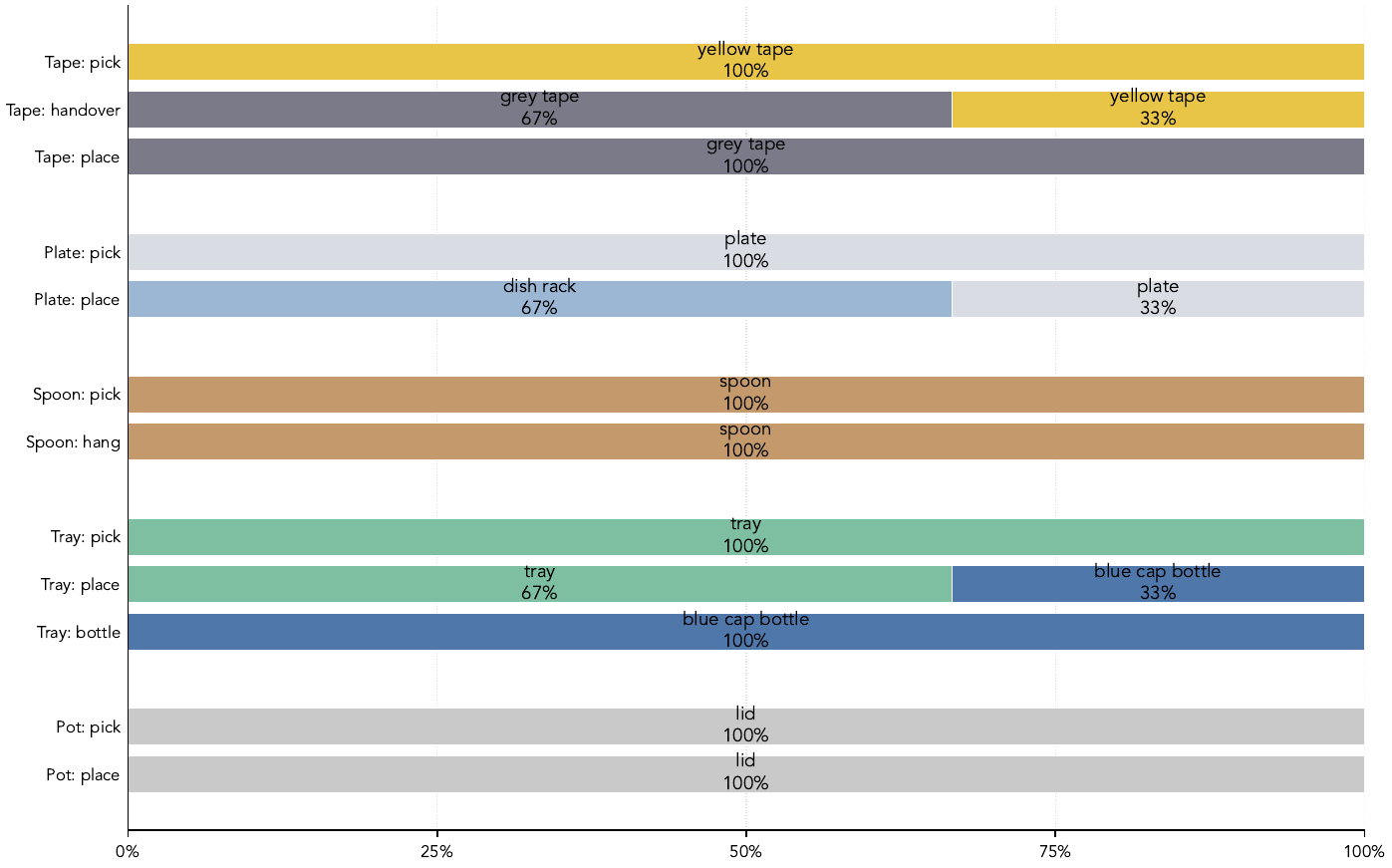}
    \caption{\textbf{Per-stage gaze selection in simulated evaluations}. For each policy trained on sim data, we compute the mean gaze pointing error to each object in the scene, and consider the object with the lowest mean error as the ``chosen'' object during that stage. \sysname{} reliably learns to pre-attend grasping, as evidenced by the fact it looks at the first object 100\% of the time. In sim, it sometimes does not look at the second stage of the task, i.e., the place action, for example in the pot and spoon hanging tasks. The tiger task is omitted as it only has one stage.}
    \label{fig:placeholder}
\end{figure}

\paragraph{\textit{How accurately can \methodname{} fixate in 3D?}} Overall, the average fixation error after gaze pre-training is 3.5\,cm computed over 800 random frames from all trajectories. Fixation quality is best seen in our included policy rollout videos, which visualize real-time fixation from the policy during execution. Despite being trained only on still frames, the policy can reliably track moving objects during robot execution.

\begin{wraptable}{r}{0.5\textwidth}
    \centering
    \small
    \label{tab:foveal_vs_periphery_avp}
     \begin{tabular}{lrr}
  \toprule
   & Periphery (cm) $\downarrow$ & Foveal (cm) $\downarrow$ \\
  \midrule
  Sim Tasks  & 4.73 & \textbf{3.51} \\
  Real Tasks & 4.44 & \textbf{3.55} \\
  \bottomrule
  \end{tabular}
    \caption{\textbf{3D fixation Euclidean error} during fixation pretraining, averaged over the final 100 training iterations of the best seed per task.}
    \end{wraptable}


\section{Model Architecture and Training Details}
Both the fixation and gripper policies use a transformer architecture which decodes actions from input image, proprioception, and goal prompt observations. The input proprioception includes the current fixation point encoded as a unit vector direction and distance. Image tokens are extracted using a frozen DINOv3~\cite{simeoni2025dinov3} ViT-S/16 backbone in a foveated manner: a pyramid of centered crops is independently passed through the encoder and concatenated along the sequence dimension, allowing the model to attend to tokens from multiple resolutions as it desires.

Each input token is given a RoPE~\cite{rope} $(x,y,t)$ position. For image tokens, $x,y$ corresponds to the location of the token within the global image, and all other tokens are given a value of the image center $(\frac{h}{2},\frac{w}{2})$. Each token is assigned $t$ corresponding to the current observation timestep, except action chunks which are assigned $t$ consistent with their future time. We use a $\theta$ value of $1,000$ for sinusoidal frequencies, lower than the default $10,000$ for RoPE to adjust for the shorter sequence length. The BC gripper policy takes in history-free single-frame observations, while the gaze policy is allowed to attend to the gaze direction of the previous 3 frames, which stabilizes gaze.

The transformer uses bi-directional attention between observation and query tokens, to improve compute efficiency over full self-attention. Each input and output modality is projected to the same token stream and decoded into outputs with a 3-layer MLP. The fixation policy is an actor-critic agent which outputs a categorical distribution over actions alongside a continuous value function estimate $v$. The target selector is a 3-layer, 256-wide MLP with GELU~\cite{gelu} activations which takes in proprioceptive input and outputs a categorical distribution over target prompts along with a critic estimate. We apply 10\% dropout to the hidden layers of the MLP to prevent overfitting to proprioceptive data, along with 10\% proprioceptive token drop during BC batch sampling. The attention and FFN hidden layers of the transformer MLPs also have 10\% dropout. Each transformer uses 6 transformer blocks, with any learnable query tokens initialized as standard normal embedding vectors with a standard deviation of 0.02 in each dimension. The token dimension of the transformer is 384, and it has an MLP expansion ratio of 4.0, with 12 attention heads. Of each head's 128 dimensions, 32 are used for temporal RoPE, 48 for $x$, and 48 for $y$.

\begin{figure}[t]
    \centering
    \includegraphics[width=\linewidth]{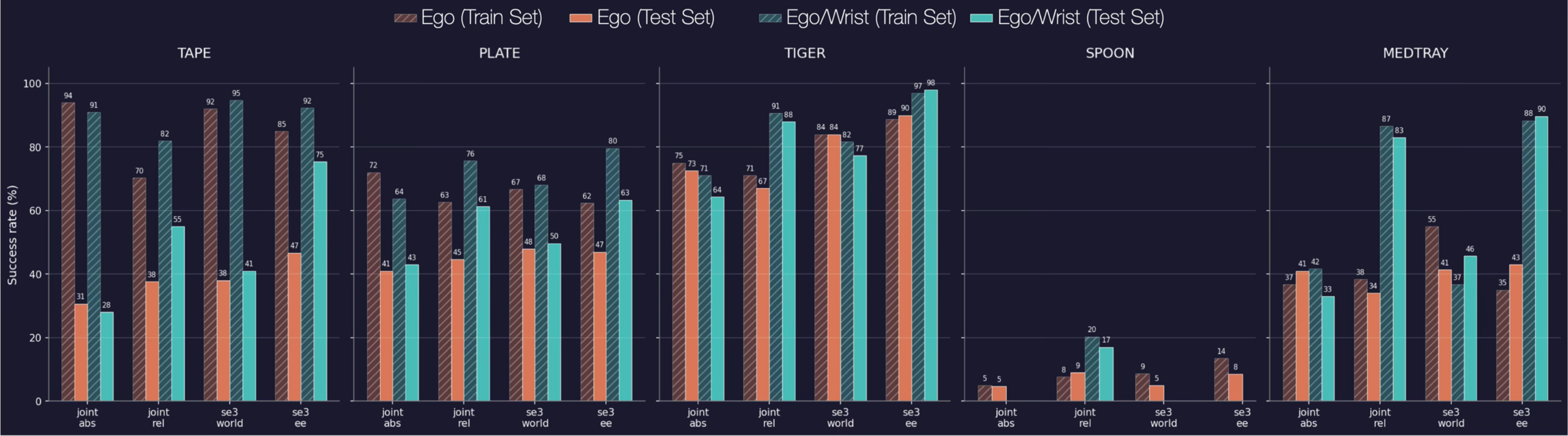}
    \caption{\textbf{Baseline action space} grid-search over all options in 5 simulated tasks. Note that faded bars are evaluated on the \textit{exact same} positional distributions as the training set to understand train-test overfitting in the simulated environments. Main-paper results are reported with the ``Test Set" distribution, which is a different random seed from teleop demonstrations. Across the board, EE-relative (rightmost in each section) outperforms other action spaces, so we use it for all real-world baselines. All baseline models exhibit signs of overfitting to the train positions on some tasks (particularly \texttt{tape} and \texttt{plate}), even for tasks where teleop positions were randomly initialized.}
    \label{fig:action_spaces}
\end{figure}

\subsection{Hyperparameters}
All training code will be released for reproducibility, but we report selected important hyperparameters here. 

\paragraph{General RL parameters}
We utilize the standard PPO update rules with value bootstrapping and discount factor $\gamma=0.95$. We utilize generalized advantage estimation~\cite{schulman2015high} with $\lambda=0.95$. We do on-policy RL updates for 10 epochs, early-terminating if KL-divergence on the discrete policy output reaches more than $0.02$. 

The gaze policy discretizes its action space and outputs a cross-product categorical distribution. Azimuth and elevation are binned into 8 polar directions, with 2 magnitudes for each direction (10$^\circ$ and 3$^\circ$) along with a zero action. Depth deltas are binned with 3 magnitudes (10\,cm, 5\,cm, 1\,cm) and a zero action. The resulting cross-producted action space has 119 bins for the policy output. We prefer discretized action spaces as they help break ties in ambiguous actions (i.e., duplicate objects).

\begin{figure}[h!]
    \centering
    \includegraphics[width=\linewidth]{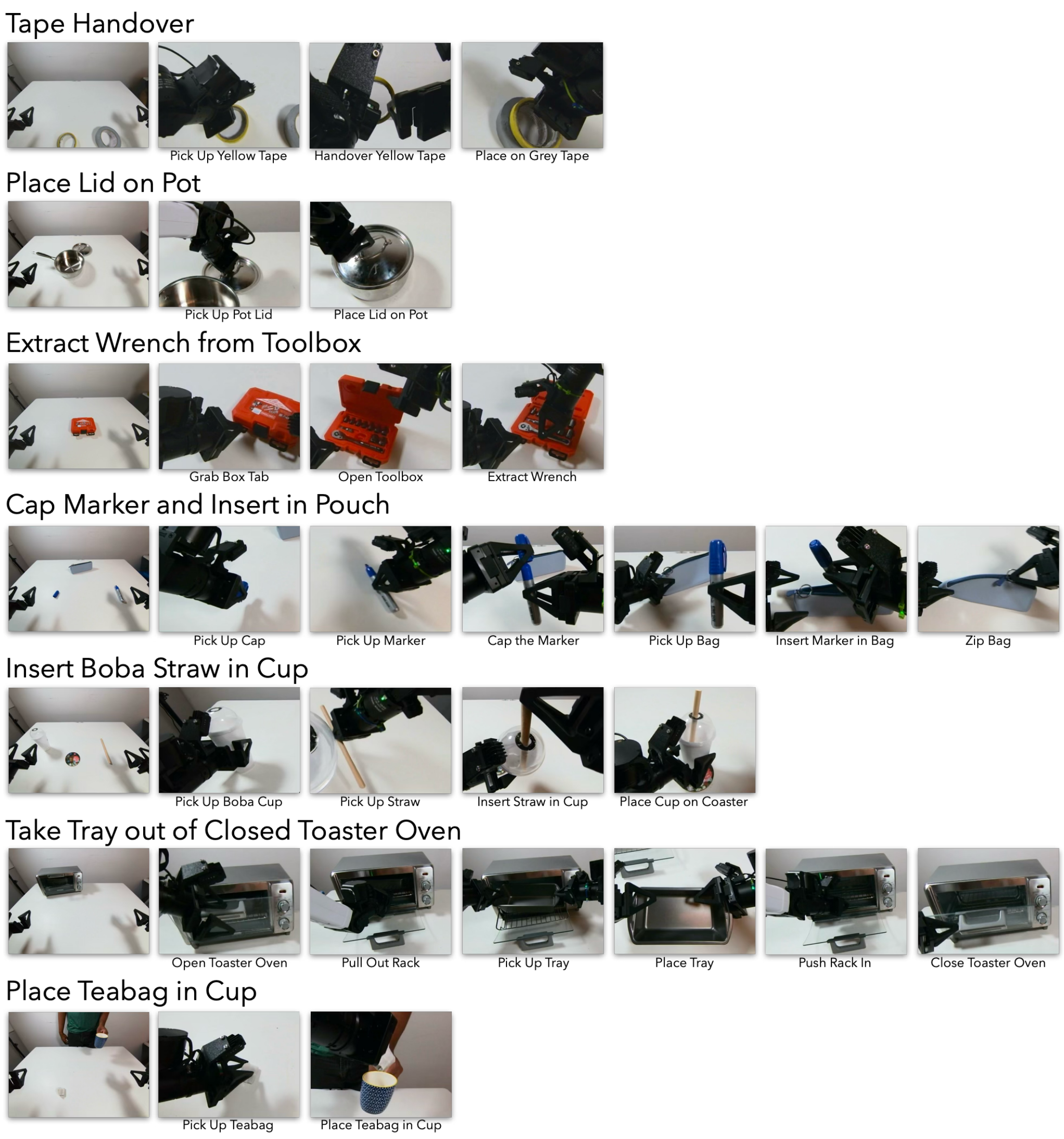}
    \caption{\textbf{Real tasks and their stages}. The tasks that we study in \sysname{} represent a suite of complex manipulation behaviors, including \textbf{bimanual coordination} (tape handover, grasping and opening a lid), \textbf{fine-grained tolerance} requirements (capping a marker, grasping a zipper, inserting a straw into a hole, inserting a marker into a pencil pouch), \textbf{challenging visual scenarios} (picking up a white teabag on a white table), and \textbf{multi-stage tasks} (6 for both marker and toaster oven tasks).}
    \label{fig:real_tasks}
\end{figure}

\begin{figure}[h!]
    \centering
    \includegraphics[width=0.85\linewidth]{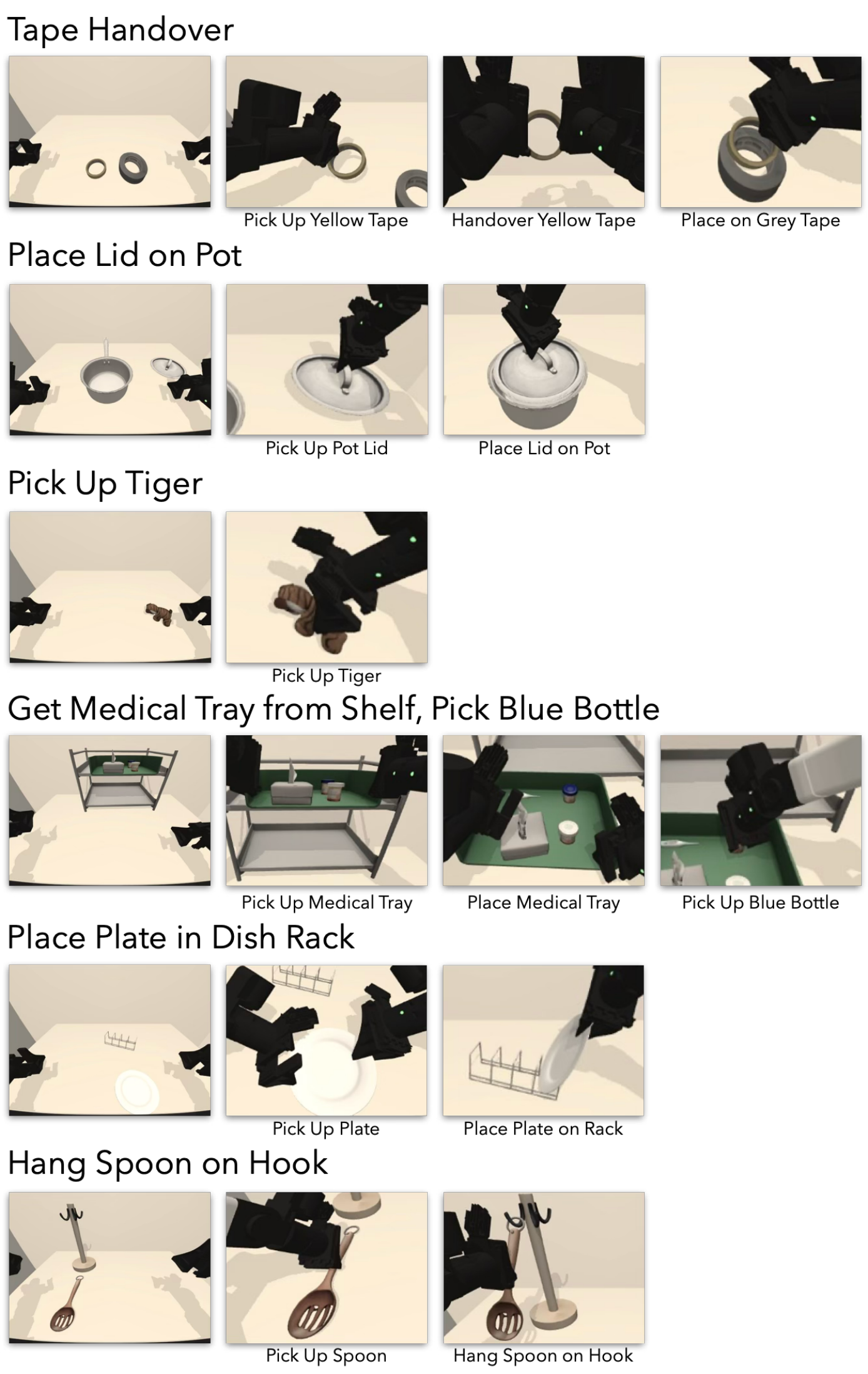}
    \caption{\textbf{Simulated tasks and their stages}. Simulated tasks represent a range of simple pick-place behaviors including single-arm (pot, spoon), single-arm-with-handover (tape), and bimanual coordination (tray, spoon). }
    \label{fig:sim_tasks}
\end{figure}

\paragraph{Gaze pre-training} Each policy rollout consists of 120 timesteps, all over a paused stereo frame. The policy is optimized with the AdamW~\cite{adamw} optimizer, with a learning rate of $10^{-4}$, $\beta_1=0.9,\beta_2=0.95$, weight decay of $0.1$, with a linear ramp over the first roughly 5\% of train steps and cosine decay over the last $30\%$. Each rollout uses 64 parallel environments, with a minibatch size of 16 environment rollouts. Policies are optimized for 15M total rollout steps, which corresponds to $1,950$ iterations, or $125,000$ unique rollouts total. We use an entropy regularization term of $0.01$ during training, which decays to $0.005$ alongside cosine decay.

\paragraph{Gaze post-training and behavior cloning}
During BC-RL training, the behavior cloning policy is completely isolated from RL gradients by having its own set of projection and decoder MLPs. Both the target selector and BC policy are optimized with Muon~\cite{jordan2024muon}, with the learning rate for the BC policy set to $5\times10^{-4}$ and the target selector $10^{-4}$. The target selector entropy is regularized with a SAC-style entropy regularizer~\cite{haarnoja2018soft} which tunes the regularization to reach a target average entropy of $70\%$ of the theoretical maximum policy entropy. During post-training, we maintain a buffer of the past 100 episodes of experience, from which we IID sample minibatches of 256 frames for the BC policy to use for SGD. We perform 4 minibatches of BC optimization for each update of the rolling experience buffer. During post-training we roll out 64 parallel environments for 30 steps with paused demonstration time, and 170 steps with time playing forward. During post-training, the video is played at 2x speed (skipping every other frame) so that each episode provides more coverage of the demonstration. In total, we roll out $10,000,000$ environment steps, which corresponds to $50,000$ unique episode rollouts, $781$ iterations, or $3,124$ BC gradient steps.

\paragraph{Training details}
Each train run uses 8 NVIDIA L40S GPUs, training with data-parallel process coordination with environments evenly split among GPUs. All reported batch sizes are for the total number, not per-GPU. Training is done in \texttt{bfloat16} mixed precision, except RoPE which is computed at \texttt{float32} resolution.

\paragraph{Baseline details} 
To decide the action space to use for all baselines, we considered the classically known options of joint-space and end-effector SE(3) space, both chunk-relative and chunk-absolute. While community wisdom has recently drifted towards increasingly using chunk-relative representations, we wanted to ensure that in our data-frugal scenario we thoroughly calibrated the performance of these. Testing such a large grid-search is infeasible in physical trials, so we ran a subset of our simulated tasks with all action spaces (Fig.~\ref{fig:action_spaces}). Across the board, we found that end-effector SE(3) relative to the first EE frame in the chunk (``EE-relative") performed the best, so we use this action space for all baseline experiments. To doubly verify this, we ran full baseline evaluations with world-relative SE(3) actions on the wrench and toaster tasks, and observed no difference in performance. Baselines are trained at 256 image resolution.

\begin{figure}[t]
    \centering
    \includegraphics[width=\linewidth]{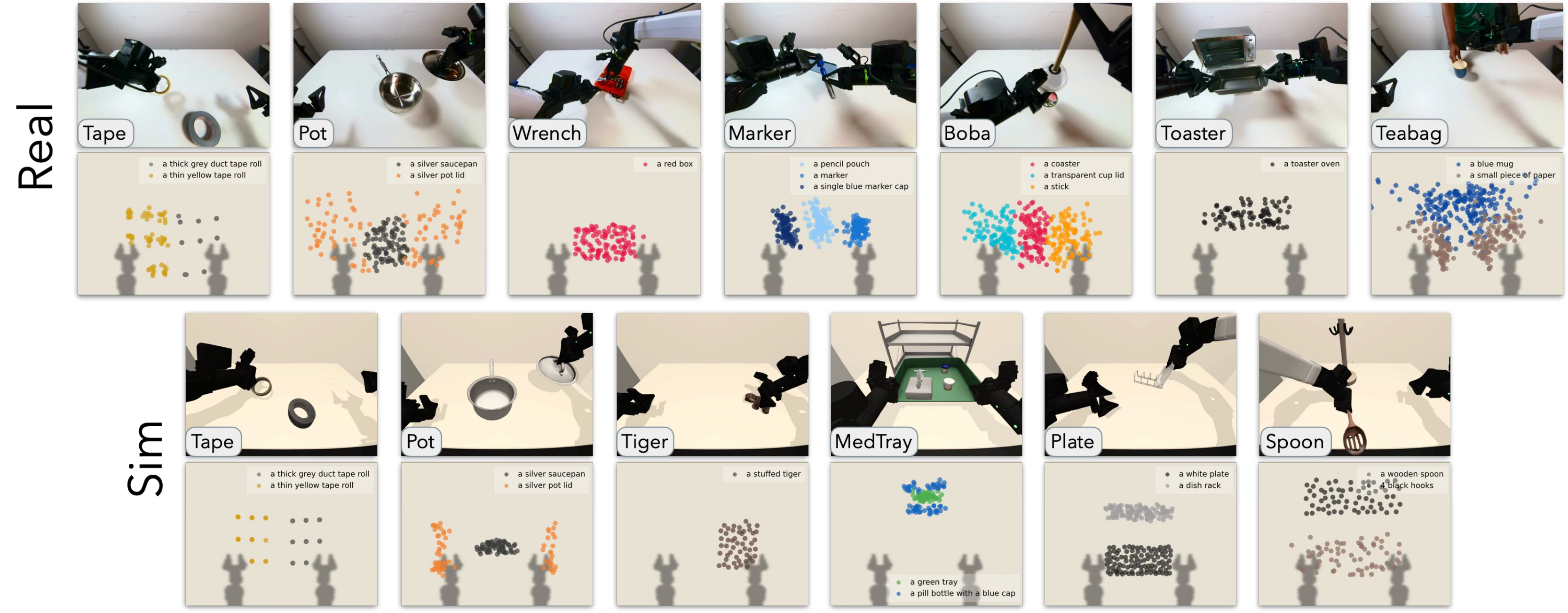}
    \caption{\textbf{Training distribution} of objects in the training set. Each dot represents an actual computed ground truth 3D location of an object which has been projected onto the table surface for visualization. Shadows of the YAM arms have been provided for scale. The distribution of objects is approximately uniformly random,  except in \texttt{tape} where the distribution is a grid. }
    \label{fig:tasks_distr}
\end{figure}

\section{Robot inference}
Action chunks are predicted as 30\,Hz open-loop trajectories, which we cubically interpolate at 200\,Hz to obtain the commanded robot positions. We apply temporal ensembling during inference with a $k$ value of 0.01.  We additionally compute the temporally-ensembled velocity across predicted action chunks and provide this as a feedforward term for the YAM follower arms, which increases fluidity of the robot movements. See included videos for unedited videos of robot execution.

All models are queried as quickly as compute will allow, with end-to-end throughput of around 40\,Hz for baselines and 20\,Hz for \methodname{} policies.

\section{Data Details}
For stage-by-stage visualizations of all task stages, see Figs.~\ref{fig:real_tasks} and~\ref{fig:sim_tasks}. Figure~\ref{fig:tasks_distr} shows the exact training distribution for all sim and real tasks, as calculated by detecting each object's starting position in 3D and plotting them projected onto the tabletop surface. As described in the Results section, notice how the \texttt{tape} task is collected in a grid, while other tasks are randomly initialized within a certain region for each object. Baseline policies are surprisingly sensitive to this positional distribution, with a large gap in performance between \methodname{} and wrist policies on this grid-collected task. This gap is lessened by collecting randomly distributed initial states.

See Tab.~\ref{tab:demo-counts} for dataset size statistics for all tasks.

\section{Simulation Details}
We use MuJoCo's Newton solver with an elliptic friction cone, implicit-fast integration at a 0.5 ms timestep, and 10 noslip iterations to ensure stable contact-rich grasping behavior in simulation. Collision meshes are generated with a convex decomposition from CoACD~\cite{wei2022coacd}, and we use the default YAM collision meshes and physical properties from MuJoCo Playground~\cite{zakka2025mujocoplayground}. Simulation environments have hand-coded initial object distributions and physical success measures, and the same parameters are used across all models. The same seed is used for environment initializations to ensure direct comparability.

\section{Evaluation Details}

\subsection{Test Distribution}
\begin{figure}[t]
    \centering
    \begin{tabular}{ccccccc}
        \includegraphics[width=0.23\linewidth]{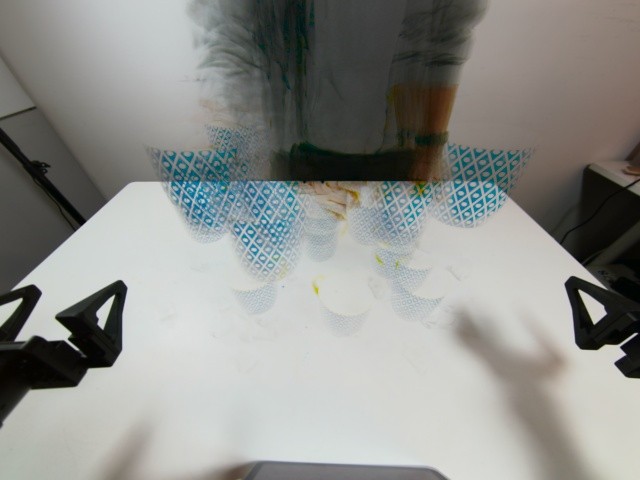} &
        \includegraphics[width=0.23\linewidth]{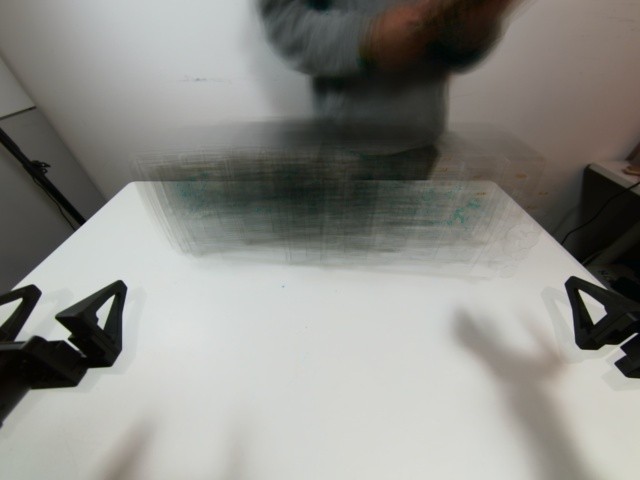} &
        \includegraphics[width=0.23\linewidth]{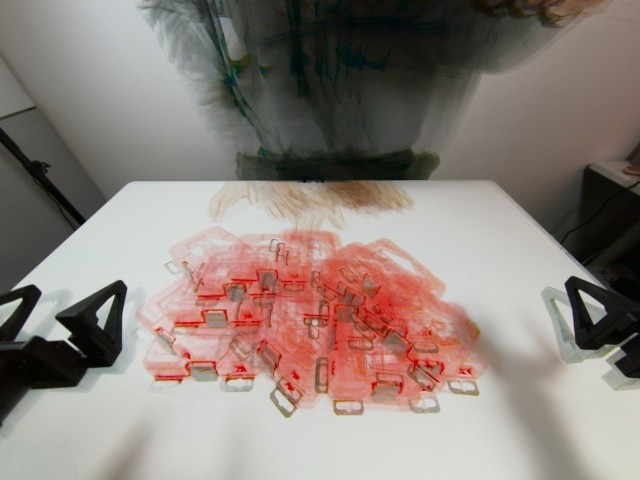} &
        \includegraphics[width=0.23\linewidth]{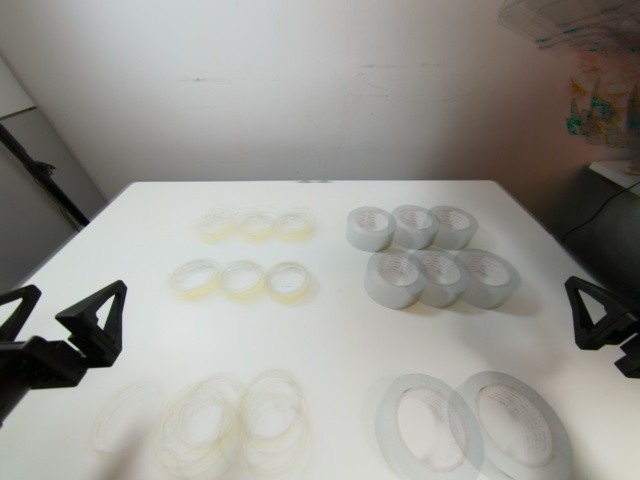} \\
        \small Tea & \small Toaster & \small Wrench & \small Tape \\[6pt]
        \multicolumn{4}{c}{%
            \includegraphics[width=0.23\linewidth]{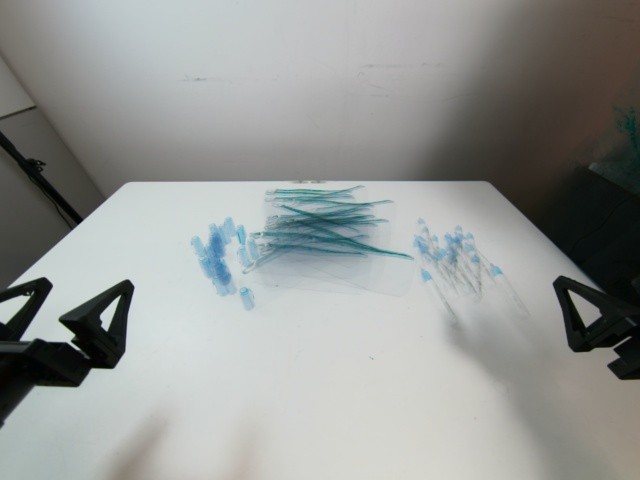}\hspace{4pt}
            \includegraphics[width=0.23\linewidth]{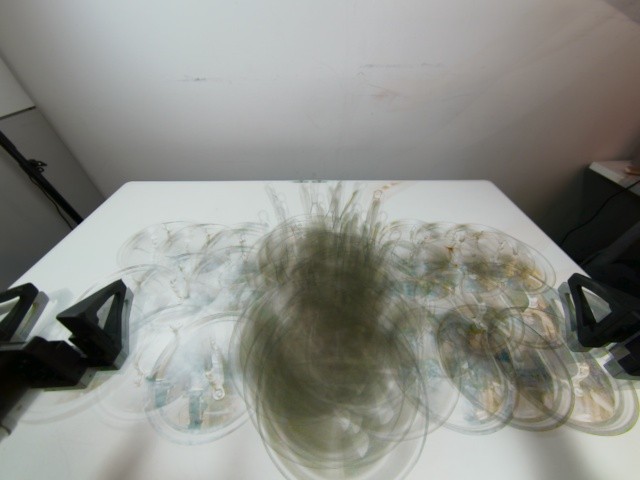}\hspace{4pt}
            \includegraphics[width=0.23\linewidth]{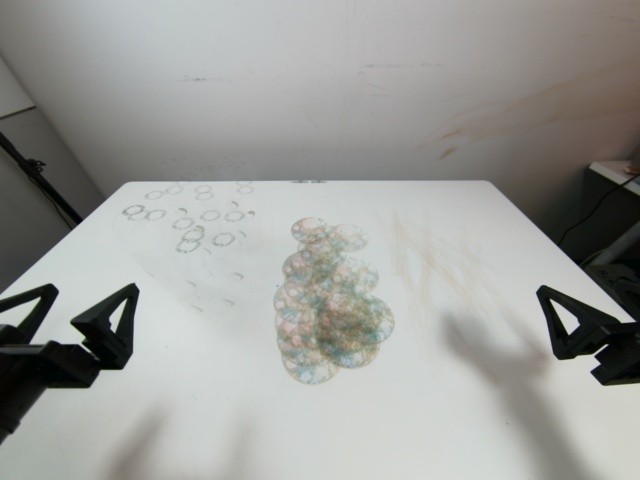}%
        } \\
        \multicolumn{4}{c}{\small Marker \hspace{0.19\linewidth} \small Pot Place \hspace{0.19\linewidth} \small Boba} \\
    \end{tabular}
    \caption{\textbf{Test distributions}, visualized by average-compositing all 25 test overlay images on top of each other. During each evaluation we standardize reset positions across policies using these.}
    \label{fig:test_distribution}
\end{figure}

See Fig.~\ref{fig:test_distribution} for a visualization of the test distributions that we use for policy evaluation. All policies are evaluated on the exact same positions, within human placement error, by aligning object positions with an overlay of the current position vs test position.

\subsection{Distractor objects}
\begin{figure}[t]
    \centering
    \begin{tabular}{ccccc}
        \includegraphics[width=0.19\linewidth]{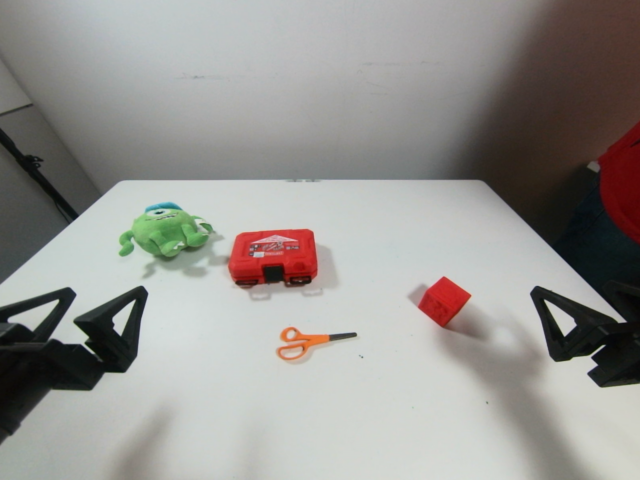} &
        \includegraphics[width=0.19\linewidth]{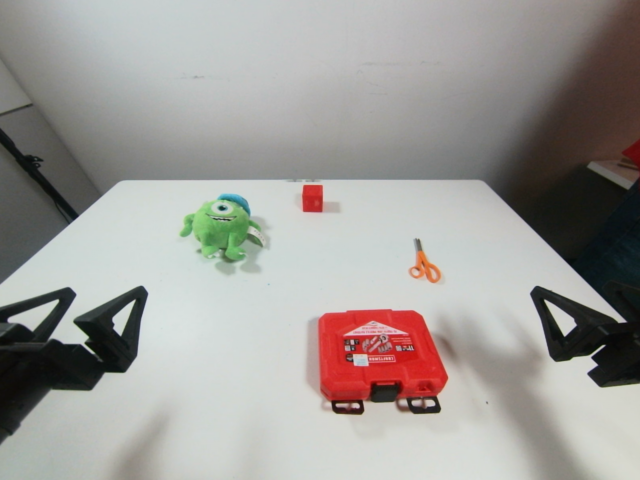} &
        \includegraphics[width=0.19\linewidth]{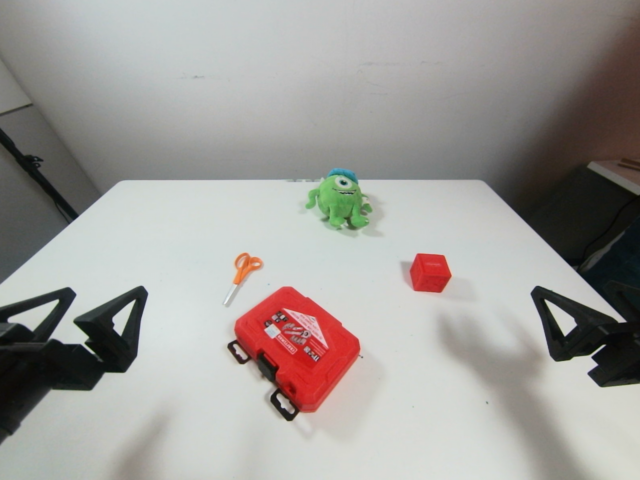} &
        \includegraphics[width=0.19\linewidth]{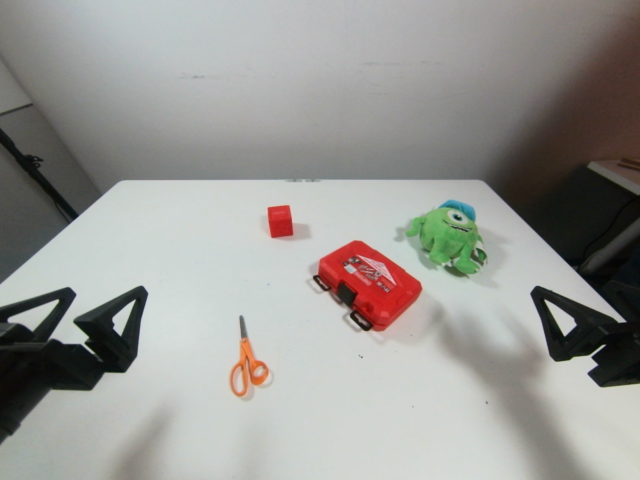} &
        \includegraphics[width=0.19\linewidth]{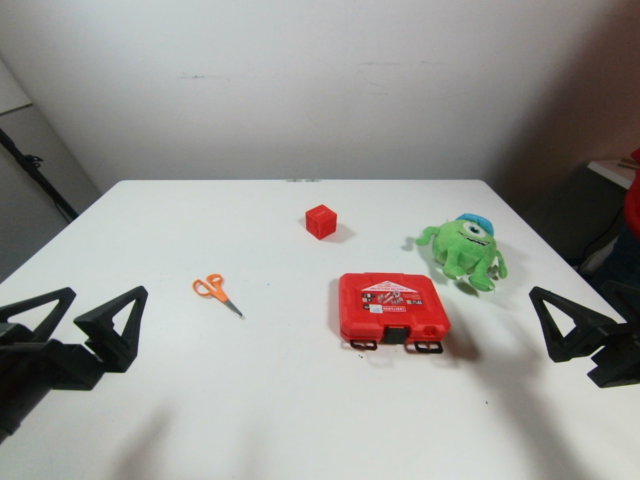} \\
    \end{tabular}
    \caption{\textbf{Subsample of distractor positions} during the wrench task (25 total).}
    \label{fig:distractors}
\end{figure}

We follow the same procedure for distractor experiments as full evals, by recording 25 positions of test objects and distractor objects and exactly repeating them for all models. See Fig.~\ref{fig:distractors} for a random subsample of initial positions of distractor objects in the \texttt{wrench} task.

\subsection{Per-stage success criteria for real tasks}
For all stages, we allow 3 retries before counting a stage as a failure, and count a failure if the robot stops moving for more than 10 seconds.

\paragraph{Tape}
\textit{Pick yellow}: the robot must grasp the yellow tape and lift it fully off the table without dropping it before the next stage. \textit{Handover}: the robot must successfully transfer the yellow tape to the right gripper without dropping it. \textit{Place}: the robot must place the yellow tape on top of the grey tape without it falling off or tilting. The static position must be stable with the center of gravity fully supported by the grey tape roll.

\paragraph{Pot}
\textit{Pick lid}: the robot must fully grasp and lift the pot lid off the table without dropping it. \textit{Place lid}: the robot must place the pot lid on the pot \textbf{with no gaps visible}. This success criterion is quite strict, as simply resting the lid on top of the pot is insufficient. The lid must fully seal the rim of the pot, which is partially why the real success of this task is significantly lower than the simulated version of the task.

\paragraph{Toaster}
\textit{Open door}: the robot must fully open the door to expose the internal oven. \textit{Pull rack}: the robot must pull out the internal wire rack from the toaster oven, and success is only marked if the rack is fully extended to reveal the pan inside, and the tray is not over-extended so far as to physically prevent pushing it back in. \textit{Pick up tray}: the robot must grab the tray inside the toaster oven with both hands (failure if only one grasps), and gently place it on the table (failure counted if the tray is dropped from more than about 3 inches or twisted during placement). \textit{Push rack in}: the robot must push the wire rack back into the oven, with failure marked if it insufficiently pushes the rack to allow door closure. \textit{Close oven door}: the robot must fully close the oven door and end without touching the door.

\paragraph{Marker}
\textit{Pick cap}: the robot must pick up the cap and reach the capping position without dropping it (mid-air drop counts as a cap failure). \textit{Pick marker}: the robot must pick up the marker from the table and reach the cap position without dropping it. \textit{Cap marker}: the robot must insert the marker into the cap such that the cap reaches a friction-fit over the marker. If the cap is partially inserted and loosely dangling, this counts as a failure. \textit{Grab bag}: the robot needs to grasp the pencil bag and place it upright close to the robot base (failure if not grasped or if the bag falls over). \textit{Insert marker}: the robot must fully insert the marker into the pencil pouch zipper enclosure, with no part of the marker sticking out. \textit{Zip bag}: the robot must fully zip the bag and let go with both grippers (zipping is considered reached when the robot fully closes the bag, with the zipper within 1cm of the edge of the bag).

\paragraph{Boba}
\textit{Pick cup}: the robot must lift the boba cup off the workspace without dropping it. \textit{Pick straw}: the robot must pick up the straw without dropping it. \textit{Insert straw}: the robot must insert the straw into the hole in the cup, with success counted if the straw is fully seated in the hole after releasing (partially inside and tilting outside is counted as a failure). \textit{Place on coaster}: the robot must place the cup on the coaster, with the cup resting flat fully supported by the coaster at the end.

\paragraph{Teabag}
\textit{Pick teabag}: the robot must grasp and lift the teabag off the table without dropping it. \textit{Place teabag in cup}: after releasing, the teabag must be fully contained inside the cup; if it hits the teacup wall and bounces out this is a failure.

\paragraph{Wrench}
\textit{Grasp box tab}: the robot must grasp the black box tab with the left hand securely without slippage during the lid movement. \textit{Open toolbox}: the toolbox lid must be fully open more than 90$^\circ$. \textit{Extract wrench}: the robot must fully grasp and lift the wrench out of the toolbox, and reach a steady state above the wrench without dropping the wrench. If the wrench falls before reaching steady state, this is considered a failure.

\begin{table}[t]
  \centering
  \resizebox{\textwidth}{!}{%
  \begin{tabular}{l rrrrrr rrrrrrrr r}
  \toprule
  & \multicolumn{6}{c}{\emph{Simulation}} & \multicolumn{7}{c}{\emph{Real}} & \\
  \cmidrule(lr){2-7} \cmidrule(lr){8-14}
  & plate & tape & tiger & hang\_spoon & medical\_tray & pot\_lid
  & tape & boba & wrench & marker\_bag & pot\_lid & tea & toaster & Total \\
  \midrule
  \# Demos & 100 & 64 & 50 & 64 & 63 & 64
           &  64 & 108 & 114 & 126 & 95 & 213 & 89 & 1278 \\
  Minutes  & 21.6 & 10.8 & 4.0 & 8.1 & 18.4 & 7.5
           &  9.5 & 31.1 & 26.3 & 53.4 & 11.3 & 26.5 & 52.0 & 287.7 \\
  \bottomrule
  \end{tabular}%
  }
  \vspace{0.5em}
  \caption{Per-task demonstration counts and total duration of demonstrations.}
  \label{tab:demo-counts}
\end{table}

\begin{table}[t]
       \centering
       \small
       \begin{tabular}{cccccc}
       \toprule
       \shortstack{\textbf{\sysname{}}\\(full)} & \shortstack{\sysname{}\\ ($-$fix. relative)
       } & \shortstack{Object gaze\\(worst obj.)} & \shortstack{Object gaze\\(best
     obj.)} & \shortstack{\sysname{}\\$-$stereo} 
     & \shortstack{\sysname{}\\$-$foveation} \\
       \midrule
       0.74 & 0.53 & 0.50& 0.69  & 0.74  & 0.72 \\
       \bottomrule
       \end{tabular}
       \vspace{0.5em}
       \caption{\textbf{Simulated Ablations}. Getting rid of fixation-relative actions (``$-$fix. relative") severely harms performance even on relatively simple pick and place tasks, dropping by 21\% overall. Restraining gaze to one constant object during training produces a 24\% drop for the worst option, and even for the best constant-gaze choice results in a 5\% reduction. Interestingly, both stereo and foveation ablations have little effect in simulated tasks, which we hypothesize is because they do not possess the visual complexity or fine-grained tolerance to require precise visual features or spatial understanding.}
       
     \end{table}

\begin{table}[t]
\centering
\small
\setlength{\tabcolsep}{7pt}
\renewcommand{\arraystretch}{1.08}
\begin{tabular}{lcccc}
\toprule
\textbf{Policy}
& \textbf{Subtask 1}
& \textbf{Subtask 2}
& \textbf{Subtask 3}
& \textbf{Success} \\
\midrule

\multicolumn{1}{l}{\textbf{wrench}}
& \footnotesize\emph{grab tab}
& \footnotesize\emph{open box}
& \footnotesize\emph{grab wrench}
& \footnotesize\emph{complete} \\
Ego-only     & 3/25  & 0/25  & 0/25 & 0/25 \\
Ego + Wrist  & 10/25 & 6/25  & 4/25 & 4/25 \\
\sysname{} & \textbf{20/25} & \textbf{7/25} & \textbf{5/25} & \textbf{5/25} \\

\cmidrule(lr){1-5}

\multicolumn{1}{l}{\textbf{tea}}
& \footnotesize\emph{grab tea}
& \footnotesize\emph{tea in cup}
& --
& \footnotesize\emph{complete} \\
Ego-only     & 3/25  & 0/25  & --   & 0/25 \\
Ego + Wrist  & 11/25 & \textbf{4/25} & -- & \textbf{4/25} \\
\sysname{} & \textbf{13/25} & \textbf{4/25} & -- & \textbf{4/25} \\

\cmidrule(lr){1-5}

\multicolumn{1}{l}{\textbf{tape}}
& \footnotesize\emph{pick yellow}
& \footnotesize\emph{handoff}
& \footnotesize\emph{place yellow}
& \footnotesize\emph{complete} \\
Ego-only     & 6/25  & 1/25  & 0/25 & 0/25 \\
Ego + Wrist  & 14/25 & 0/25  & 0/25 & 0/25 \\
\sysname{} & \textbf{23/25} & \textbf{12/25} & \textbf{8/25} & \textbf{8/25} \\

\bottomrule
\end{tabular}
\vspace{0.5em}
\caption{\textbf{Task success with distractors.} Each cell reports the number of trials out of 25 reaching the corresponding subtask. Success denotes end-to-end task completion. Bold indicates the best policy for each task-subtask.}
\label{tab:distractor_success_breakdown}
\end{table}
\begin{table}[h]
  \centering
  \small
  \setlength{\tabcolsep}{4pt}
  \label{tab:sim_hero}
  \begin{tabular}{l ccc ccc ccc}
  \toprule
  Policy & \multicolumn{3}{c}{Plate} & \multicolumn{3}{c}{Tape} & \multicolumn{3}{c}{Tiger} \\
  \cmidrule(lr){2-4} \cmidrule(lr){5-7} \cmidrule(lr){8-10}
   & med & mn$\pm$sd & max & med & mn$\pm$sd & max & med & mn$\pm$sd & max \\
  \midrule
  BC ego-mono & 54 & 52$\pm$5 & 55 & 38 & 38$\pm$5 & 41 & 76 & 78$\pm$5 & 83 \\
  BC ego-stereo & 41 & 41$\pm$1 & 41 & 52 & 52$\pm$4 & 55 & 86 & 86$\pm$2 & 87 \\
  BC ego+wrist & 71 & 71$\pm$3 & 74 & 79 & 79$\pm$3 & 81 & 96 & 95$\pm$2 & 96 \\
  \sysname{} & 67 & 67$\pm$3 & 70 & 75 & 76$\pm$2 & 79 & 97 & 97$\pm$1 & 98 \\
  \bottomrule
  \end{tabular}
  \\[1.5ex]
  \begin{tabular}{l ccc ccc ccc}
  \toprule
  Policy & \multicolumn{3}{c}{Spoon} & \multicolumn{3}{c}{Tray} & \multicolumn{3}{c}{Pot} \\
  \cmidrule(lr){2-4} \cmidrule(lr){5-7} \cmidrule(lr){8-10}
   & med & mn$\pm$sd & max & med & mn$\pm$sd & max & med & mn$\pm$sd & max \\
  \midrule
  BC ego-mono & 22 & 21$\pm$5 & 26 & 50 & 48$\pm$3 & 50 & 69 & 69$\pm$1 & 70 \\
  BC ego-stereo & 23 & 24$\pm$2 & 26 & 40 & 39$\pm$6 & 44 & 62 & 64$\pm$5 & 69 \\
  BC ego+wrist & 27 & 27$\pm$1 & 27 & 85 & 83$\pm$3 & 85 & 89 & 89$\pm$1 & 90 \\
  \sysname{} & 26 & 27$\pm$2 & 30 & 77 & 76$\pm$2 & 78 & 90 & 89$\pm$3 & 91 \\
  \bottomrule
  \end{tabular}
  \vspace{1em}
  \caption{\textbf{Full sim success statistics (\%).} Median, mean, and std are reported over 3 seeds for each policy/task combination. Each evaluation of a seed is done over 100 trials.}
\end{table}

\clearpage

\clearpage
\bibliography{example}  %

\clearpage

\end{document}